\documentclass[10pt,twocolumn,letterpaper]{article}

\usepackage{cvpr}              % To produce the CAMERA-READY version
\usepackage{graphicx}
\usepackage{amsmath}
\usepackage{amssymb}
\usepackage{booktabs}
\usepackage{listings}
\definecolor{codegray}{gray}{0.95}

\definecolor{mygray}{gray}{0.9}
\definecolor{eblue}{HTML}{1E90FF}
\definecolor{myblue}{HTML}{8080FF}
\definecolor{myorange}{HTML}{FF9461}
\definecolor{mygreen}{HTML}{228B22}
\definecolor{mysalmon}{HTML}{E9967A}
\usepackage{amsthm}

\newtheorem{proposition}{Proposition}

\usepackage[dvipsnames]{xcolor}
\newcommand{\sr}[1]{\small{(\textcolor{myblue}{$\uparrow$#1})}}

\newcommand{\sg}[1]{\small{(\textcolor{myblue}{$\downarrow$#1})}}

\DeclareUnicodeCharacter{2212}{\ensuremath{-}}

\DeclareMathOperator{\softmax}{Softmax}

\newcommand{\MSP}{MSP~\cite{DanMSP}\xspace}
\newcommand{\MSPTAG}{MSP*}

\newcommand{\GENTAG}{GEN*}

\newcommand{\ReActTAG}{ReAct*}
\newcommand{\KNNTAG}{KNN*}
\newcommand{\SCALETAG}{SCALE*}
\newcommand{\fDBDTAG}{fDBD*}

\newcommand{\NPOS}{NPOS~\cite{npos}\xspace}

\newcommand{\ReAct}{ReAct~\cite{ReAct}\xspace}

\newcommand{\SCALE}{SCALE~\cite{xu2024scaling}\xspace}
\newcommand{\KNN}
{KNN~\cite{sun2022knnood}\xspace}

\newcommand{\CIDER}{CIDER~\cite{ming2022cider}\xspace}

\newcommand{\GEN}{GEN~\cite{xixi2023GEN}\xspace}
\newcommand{\fDBD}{fDBD~\cite{liu2023fast}\xspace}
 
\usepackage{graphicx}
\usepackage{amsmath}
\usepackage{amssymb}
\usepackage{booktabs}
\usepackage{multirow}

\usepackage{multicol}
\usepackage{comment}
\usepackage{adjustbox}
\usepackage{makecell}
\usepackage{mathtools}
\usepackage{pifont}
\usepackage{booktabs,tabularx, colortbl}
 
\definecolor{TableGray}{gray}{0.9}

\newcommand{\doublerule}{\midrule\midrule}

\definecolor{commentcolor}{RGB}{110,134,185}   % define comment color
\def\id{\mathcal{D}_{\text{ID}}}
\def\idlabel{\mathcal{Y}_{\text{ID}}}

\def\ood{\mathcal{D}_{\text{OOD}}}

\definecolor{cvprblue}{rgb}{0.21,0.49,0.74}
\usepackage[pagebackref,breaklinks,colorlinks,allcolors=cvprblue]{hyperref}

\def\paperID{6482} % *** Enter the Paper ID here
\def\confName{CVPR}
\def\confYear{2026}

\title{Enhancing Out-of-Distribution Detection with Extended Logit Normalization}

\author{Yifan Ding\textsuperscript{1} \and Xixi Liu\textsuperscript{2} \and Jonas Unger\textsuperscript{1} \and Gabriel Eilertsen\textsuperscript{1} \\
\textsuperscript{1}Linköping University \quad \textsuperscript{2}Imperial College London\\
{\tt\small \textsuperscript{1}\{yifan.ding, jonus.unger, gabriel.eilertsen\}@liu.se \quad \textsuperscript{2}x.liu2@imperial.ac.uk}
}

\begin{document}
\maketitle
\begin{abstract}

\noindent Out-of-distribution (OOD) detection is essential for the safe deployment of machine learning models. Extensive work has focused on devising various scoring functions for detecting OOD samples, while only a few studies focus on training neural networks using certain model calibration objectives, which often lead to a compromise in predictive accuracy and support only limited choices of scoring functions. In this work, we first identify the feature collapse phenomena in Logit Normalization (LogitNorm), then propose a novel hyperparameter-free formulation that significantly benefits a wide range of post-hoc detection methods. To be specific, we devise a feature distance-awareness loss term in addition to LogitNorm, termed \textbf{ELogitNorm}, which enables improved OOD detection and in-distribution (ID) confidence calibration. Extensive experiments across standard benchmarks demonstrate that our approach outperforms state-of-the-art training-time methods in OOD detection while maintaining strong ID classification accuracy. Our code is available on: \url{https://github.com/limchaos/ElogitNorm}.

\begin{figure}[t]
  \centering
  \includegraphics[width=1\linewidth]{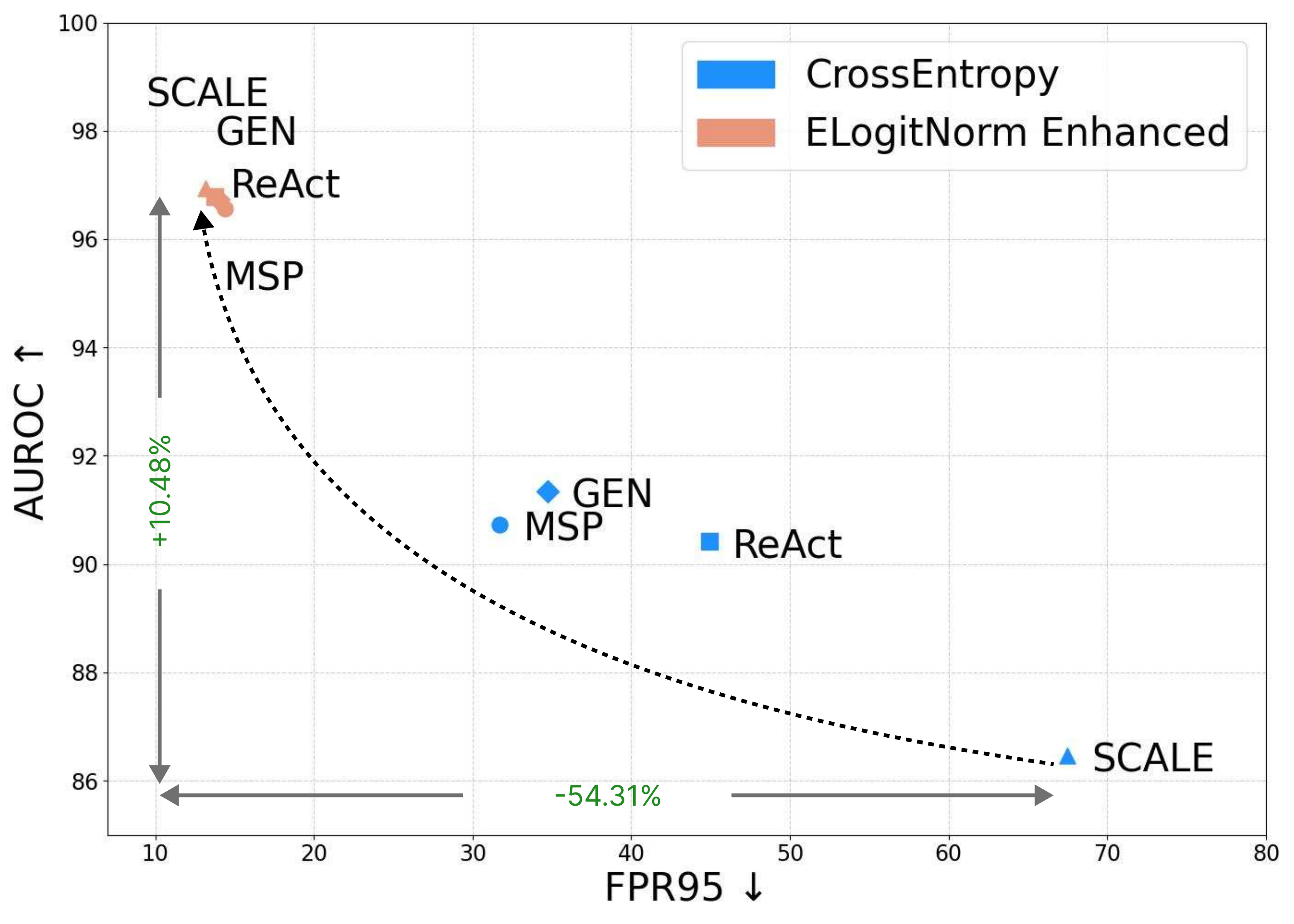}

   \caption{\emph{Effectiveness of \textbf{ELogitNorm} applied with 4 Baseline OOD Methods}. The OOD performance,  as measured by AUROC $\%$ (higher is better) and FPR95 $\%$ (lower is better). The in-distribution dataset is CIFAR-10 and the scores have been averaged over four far-OOD datasets in the OpenOOD benchmark~\cite{zhang2023openood}.}
   \label{fig:onecol}
\end{figure}

%%%%% Rerevised by Xixi
\iffalse
Out-of-distribution (OOD) detection is essential for the safe deployment of deep models. Extensive works have focused on devising various scoring functions 
%based on the information from feature, logit and probability space of
given a pretrained classier, which is typically trained with cross-entropy loss. Few works focus on mitigating the inherent overconfidence of neural network due to CE loss via logit normalization or deep metric learning with a comprise of predictive accuracy and limited choices of scoring functions. In this work, we first identify the feature collapse phenomena in LogitNorm, then propose a novel hyperparameter-free formulation that significantly benefits a wide range of post-hoc detection methods. To be specific, we devise a feature distance-awareness loss term in addition to LogitNorm, termed \textbf{ELogitNorm}, which enables improved OOD detection and in-distribution (ID) confidence calibration. Extensive experiments across standard benchmarks demonstrate that our approach outperforms state-of-the-art training-time methods in OOD detection while maintaining strong ID classification accuracy. 
%\textcolor{red}{add some numbers here}
\fi

\end{abstract}
\section{Introduction}
\label{sec:introduction}

The reliability of learning-based systems is a cornerstone for their successful deployment in safety critical and socially impactful applications. Despite their remarkable performance, deep networks often assume that the training and test data share the same underlying distribution. In real world scenarios, however, this assumption rarely holds true. Models frequently encounter samples that differ from the training distribution, leading to unreliable predictions and degraded performance. 
%% added by Xixi
%Such distribution shift can be roughly categorized into covariate shift (\eg, test samples share the same label space but with different styles or backgrounds compared to training samples), and semantic shift (\eg, test samples does not share overlapped label space while similar styles). In this work, we focus on detecting samples with semantic shift. 
%Out-of-distribution (OOD) detection aims to identify samples that are semantically different from in-distribution (ID) data, and a 
Thus, a large body of research has been dedicated to developing methods for solving this problem by means of out-of-distribution (OOD) detection~\cite{DanMSP, KL-matching, maha, energybased_ood, Generalized_ODIN, xixi2023GEN, ReAct, xu2024scaling, ash, logitnorm, cider, du2022vos}. Many previous works develop \emph{post-hoc} OOD scoring methods using information from the feature space, logit space, or probability space of the trained model~\cite{DanMSP, KL-matching, energybased_ood,xixi2023GEN}. Some approaches require access to training data statistics~\cite{KL-matching, maha, ViM}, while others focus on reshaping feature representations to enhance the separation between in-distribution (ID) and OOD features~\cite{ReAct, rank, xu2024scaling, ash}. Most of these methods assume that the classifier is trained solely with Cross-Entropy loss, yet their performance remains suboptimal. To address these limitations, several studies have proposed synthesizing outlier features and images from ID features~\cite{npos, du2022vos, cider, du2023dream, logitnorm}, or reformulating classification as a deep metric learning task~\cite{cider}. While these approaches improve OOD detection compared to post-hoc methods, they often rely on external generative models such as Stable Diffusion~\cite{stable_diffusion}, require two-stage training~\cite{npos, du2022vos, du2023dream}, impose restrictions on OOD scoring methods, or suffer from a trade-off between classification accuracy and OOD detection performance.

%%% To be continued.
%Logit normalization (LogitNorm)~\cite{logitnorm} enforces a constant norm on the logit vector, which can boost the performance of post-hoc OOD detection methods. However, as we will show in this paper, LogitNorm also leads to a certain amount of collapse of the features learned by the neural network, leading to suboptimal OOD detection performance.

%% added by Xixi
\noindent LogitNorm~\cite{logitnorm} dives into the training dynamics of cross-entropy loss and enforces a normalization on the logit vector to mitigate the overconfidence of predictions. However, it improves OOD performance as a compromise of ID classification performance and limited post-hoc scoring functions. In this work, we diagnose the limitations of LogitNorm and reveal that the learned features tend to collapse toward the origin while being compressed into a few dominant directions. To address this issue, we introduce a novel training objective that accounts for the distances to class-specific decision boundaries within LogitNorm. We refer to this approach as \emph{Extended Logit Normalization} (\textbf{ELogitNorm}) and demonstrate that it effectively prevents feature collapse while improving OOD detection performance. Moreover, models trained with our proposed loss can be seamlessly integrated with most post-hoc OOD scoring functions, whereas LogitNorm may exhibit degraded OOD performance under certain post-hoc scoring methods (see Fig.~\ref{fig:ex}). In addition, our method achieves better confidence calibration (see Tab.~\ref{tab:ece_scores}).  It has also been consistently observed that no single method dominates across all benchmarks~\cite{zhang2023openood}. For instance, approaches such as \SCALE perform strongly on ImageNet-1K but struggle on CIFAR-10. As shown in Fig.~\ref{fig:onecol}, \SCALE is outperformed by alternative methods. After applying calibrated training with our proposed \textbf{ELogitNorm}, all post-hoc OOD detection methods cluster around a similar level of OOD detection accuracy.

%%%%%%%%%%%%%%

%\paragraph{Contributions}

% enhancing OOD performance across a wide range of OOD scoring methods, see 
% In this work, we devise a novel loss based on the decision boundary among each class, in addition to the cross-entropy loss, to train a classifier. 

% \begin{itemize}
%     \item The novel classifier formulation enhances the OOD performance across different post-hoc OOD scoring methods (\ie,  MSP~\cite{DanMSP}, Max-Logit~\cite{KL-matching}, and GEN~\cite{xixi2023GEN}), and enhances methods such as ReAct~\cite{ReAct}.
%     \item Compared to training loss modification methods, our methods are compatible with different OOD scores with improved performance while maintaining the classification accuracy.
%     \item Our method is effective and hyperparameter-free, while achieves superior OOD performance on both near-OOD and far-OOD benchmarks. Interestingly, our method also leads to better-calibrated classifiers in terms of expected calibration error (ECE).
% \end{itemize}

\vspace{2mm}
\noindent In summary, our approach offers several key contributions and advantages:
\begin{itemize}
\item We identify a feature collapse phenomenon in LogitNorm, which restricts its applicability to a wider range of post-hoc OOD scoring functions and leads to reduced classification accuracy.
\item \textbf{ElogitNorm} improves OOD detection performance across diverse post-hoc OOD scoring methods. In contrast to existing training approaches, our method maintains classification accuracy while ensuring broader compatibility with various OOD scoring functions, cf. Fig.~\ref{tab:elogitnorm_results}.
\item The proposed method is simple, hyperparameter-free, and consistently achieves superior OOD performance on both near-OOD and far-OOD benchmarks. Moreover, it produces better-calibrated classifiers, as reflected by lower expected calibration error (ECE) values, cf. Table~\ref{tab:ece_scores}.
\end{itemize}

 % added by Xixi
%  It is hyperparameter-free and achieves superior OOD performance on both near-OOD and far-OOD benchmarks.  
%  maintain predictive acccuracy and better calibrated in terms of ECE.
%  Is is compatible with various OOD scoring methods and enhances the OOD performance compared to CE loss.

% \begin{figure}[t]
%   \centering
%   %\fbox{\rule{0pt}{2in} \rule{0.8\linewidth}{0pt}}
%   \includegraphics[width=0.75\linewidth]{figs/intro.pdf}

%    \caption{Distance to origin \(||\mathbf{z}||\) and distance to decision boundary \(\mathcal{D}(\mathbf{z}, b)\), where \(b\) is the corresponding decision boundary.}
%    \label{fig:prop}
% \end{figure}

\section{Preliminaries}
\label{sec:preliminaries}

%-------------------------------------------------------------------------

\subsection{Out-of-distribution Detection}

In image classification, ID samples are defined by a fixed set of semantic categories \( \idlabel \) with joint distribution \( \id \), where \( \forall(\mathbf{x}, y) \sim \id, \; y \in \idlabel \). In open-world scenarios, additional unseen categories naturally emerge, forming the OOD space \( \ood \). In this setting, OOD detection aims to achieve two primary objectives \cite{zhang2023openood}.
The first is to develop a discriminative model that accurately classifies ID samples drawn from \( \id \). The second objective is to construct a confidence score \( S \) that effectively distinguishes between ID and OOD samples. Formally, given an input sample \( x \) and a neural network parametrized by $\theta$, the OOD detection is defined as follows:
\begin{equation}
    x \in
    \begin{cases}
      \mathcal{X}_{\text{ID}}, & \text{if \( S_\theta(x) \geq \lambda \)} \\
      \mathcal{X}_{\text{OOD}}, & \text{if \( S _\theta(x) < \lambda \)} ,
    \end{cases}       
\end{equation}
where \( \lambda \) is a predefined threshold, and $S_\theta(\cdot)$ is an OOD scoring function such as ~\cite{DanMSP, KL-matching, energybased_ood, xixi2023GEN}.

\subsection{Logit Normalization}

\label{subsec: LN}
% revised by Xixi
%Given a classifier trained with cross-entropy loss, raw logits are typically unbounded, which can lead to overconfident predictions and suboptimal generalization, particularly in the presence of OOD samples. Logit Normalization (LogitNorm)~\cite{logitnorm} aim to mitigate this inherent overconfidence by ensuring that the logits maintain a controlled magnitude during training, improving robustness and calibration. Mathematically, let \( \mathbf{f} = f(\mathbf{x}; \theta) \) be the logit vector corresponding to an input \( \mathbf{x} \) and neural network $f$ parametrized by \( \theta \).  

Logit Normalization (LogitNorm)~\cite{logitnorm} is a technique used to stabilize the output logits of a neural network by constraining their norm. In classification models, raw logits are typically unbounded, which can lead to overconfident predictions and suboptimal generalization, particularly in the presence of OOD samples. To address this, LogitNorm ensures that the logits maintain a controlled magnitude, improving robustness and calibration. Mathematically, let \( \mathbf{f} = f(\mathbf{x}; \theta) \) be the logit vector corresponding to an input \( \mathbf{x} \) and neural network $f$ parametrized by \( \theta \). LogitNorm normalizes the logits to produce a unit logit vector:
\begin{equation}
\tilde{\mathbf{f}} = \frac{\mathbf{f}}{|| \mathbf{f} ||},
\end{equation}
where \( || \mathbf{f} || \) denotes the \(\ell_2\)-norm of the logit vector. This operation projects the logits onto a unit sphere, preventing excessive scaling and enforcing a uniform logit distribution. The corresponding loss function for LogitNorm is given by:
\begin{equation}
\mathcal{L}_{\text{LogitNorm}}(f(\mathbf{x}; \theta), y) = - \log \frac{e^{f_y/\tau || \mathbf{f} ||}}{\sum_{i=1}^{c} e^{f_i/\tau || \mathbf{f} ||}},
\label{eqn:logitnorm}
\end{equation}
where \( c \) is the number of classes, \( f_y \) represents the logit corresponding to the correct class, and \( \tau \) is a temperature hyperparameter. Replacing the standard cross-entropy objective, the LogitNorm training objective is optimizing:

\begin{equation}
\min_{\theta} \mathbb{E}_{\mathcal{P}_{\mathcal{X}\mathcal{Y}}} \mathcal{L}_{\text{LogitNorm}}(f(\mathbf{x} ; \theta), y).
\end{equation}

\section{Method: Extended Logit Normalization}
\label{sec:method}

\begin{figure*}[ht!]
  \centering

  \begin{subfigure}{0.23\linewidth}
    \includegraphics[width=1\linewidth]{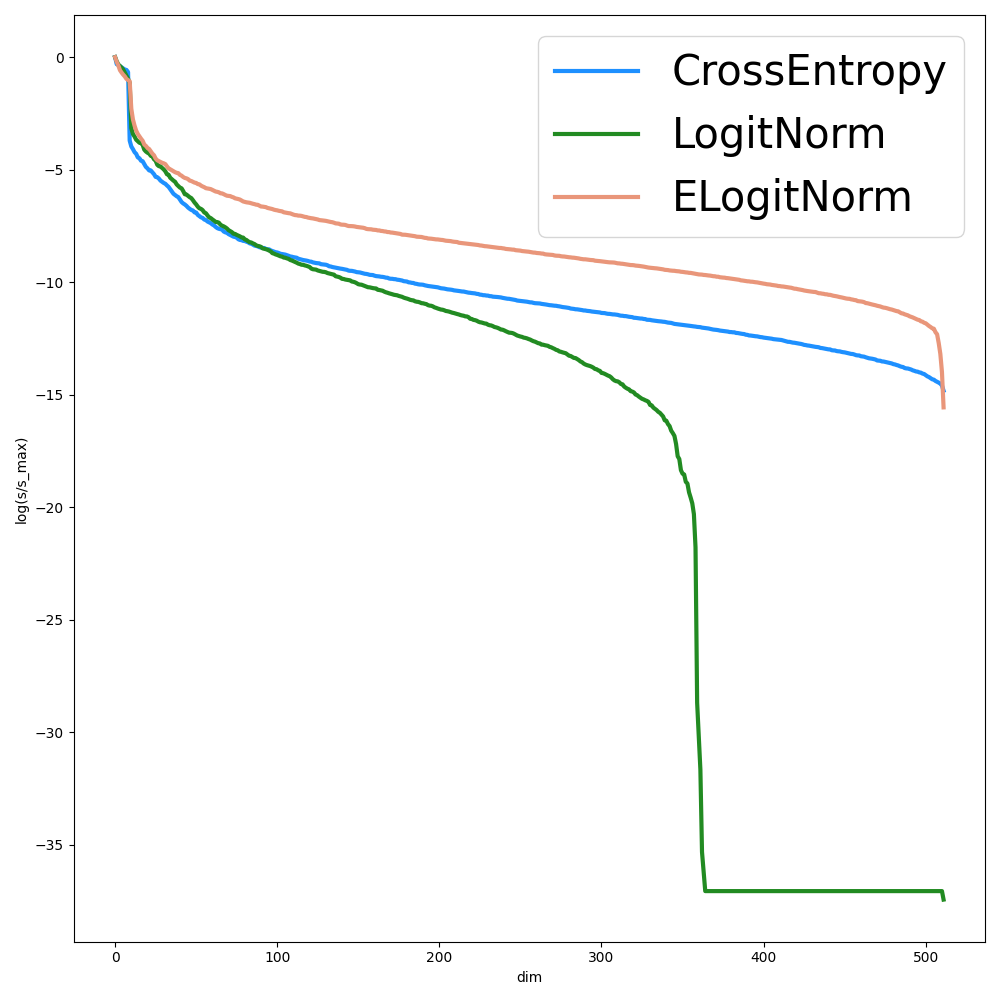}
    \caption{Singular value spectrum $\Sigma_\text{cov}^{\text{ID}}$}
    \label{fig:motivation-b}
  \end{subfigure}
    \begin{subfigure}{0.23\linewidth}
    \includegraphics[width=1\linewidth]{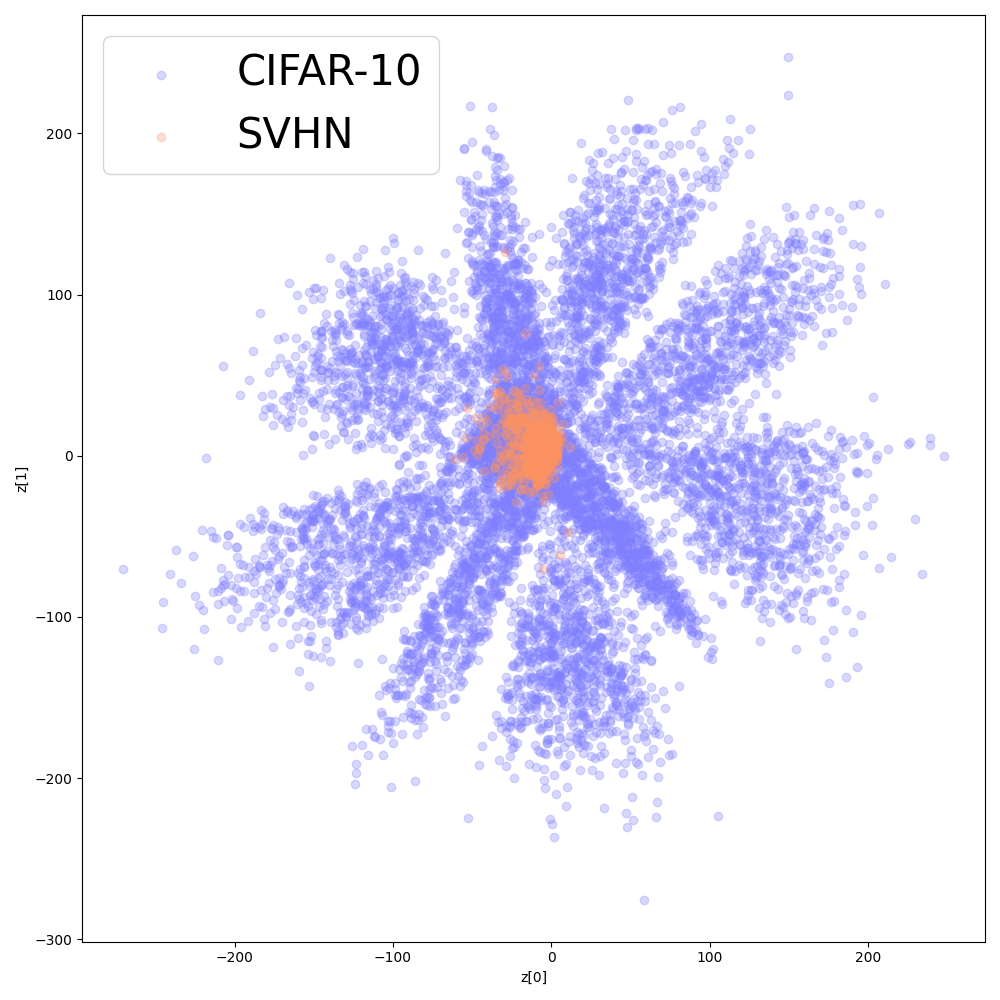}
    \caption{CIFAR-10 v.s. SVHN}
    \label{fig:motivation-a}
  \end{subfigure}
    \begin{subfigure}{0.23\linewidth}
    \includegraphics[width=1\linewidth]{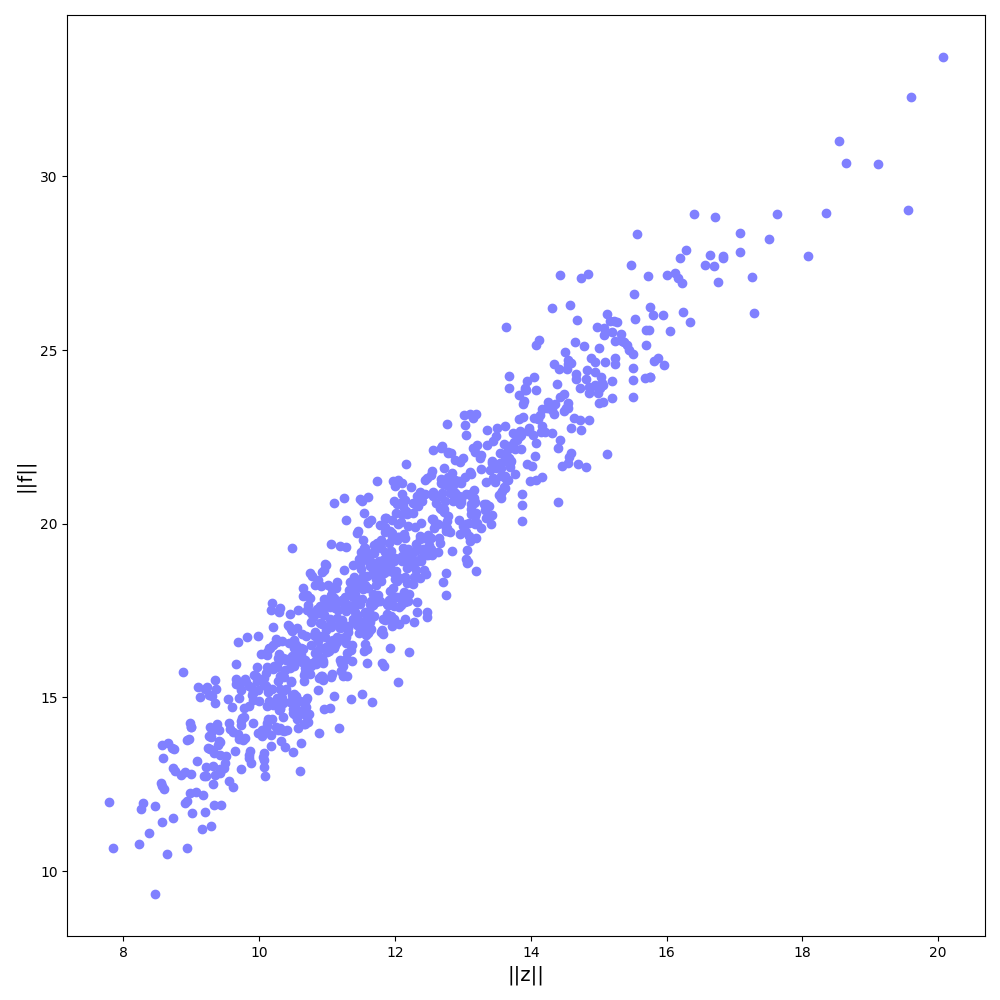}
    \caption{\(||\mathbf{f}||\) v.s. \(||\mathbf{z}||\)}
    \label{fig:motivation-c}
  \end{subfigure}
  \begin{subfigure}{0.23\linewidth}
    \includegraphics[width=1\linewidth]{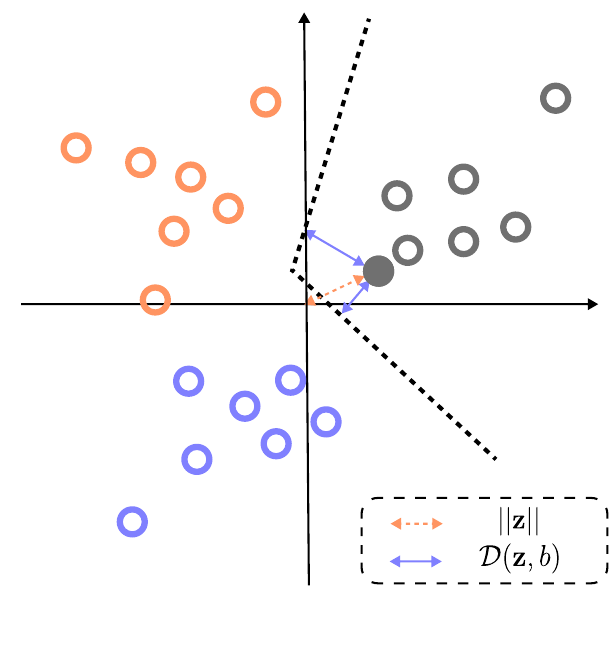}
    \caption{\(\mathcal{D}(\mathbf{z}, b)\) v.s. \(||\mathbf{z}||\)}
    \label{fig:motivation-d}
  \end{subfigure}

  \caption{(a): The singular value spectrum (log scale for better visualization) of the covariance of training ID embedding \(\mathbf{z}\) under different training losses including \textcolor{eblue}{Cross-Entropy}, \textcolor{mygreen}{LogitNorm}~\cite{logitnorm}, and \textcolor{mysalmon}{\textbf{ELogitNorm} (Ours)}. The ID data is CIFAR-10 trained on ResNet-18. (b): We train a ResNet18 on CIFAR-10 and set feature \(\mathbf{z \in \mathbb{R}^2}\) before the last penultimate layer for visualization. (c) $||\mathbf{z}||$ and $||\mathbf{f}||$ denote the feature norm and logit norm, respectively. (d): Distance to origin \(||\mathbf{z}||\) and distance to decision boundary \(\mathcal{D}(\mathbf{z}, b)\), where \(b\) is the corresponding decision boundary.  }
  \label{fig:motivation}
\end{figure*}
In the following, we first identify two types of feature collapse in LogitNorm, and then demonstrate that they can be overcome by our proposed method.

\paragraph{Notation.} To analyze the limitations of LogitNorm, we consider the feature space of the classifier. To this end, let \( \mathbf{f} = \mathbf{W}^\top \mathbf{z} + \mathbf{b} \) be the logits of a neural network classifier, where \( \mathbf{W} \in \mathbb{R}^{m \times c} \) is the weight matrix, \( \mathbf{z} \in \mathbb{R}^m \) is the feature vector before the final layer, and \( \mathbf{b} \in \mathbb{R}^c \) is the bias term.  

\subsection{Motivating Observation}

Although LogitNorm aims to mitigate overconfidence by enforcing normalization on logits $\mathbf{f}$ during training, we find that it inadvertently induces feature dimension collapse. In addition, even when trained solely with cross-entropy loss, OOD feature representations tend to cluster near the origin rather than being distributed across other low-likelihood regions (\eg decision boundaries). We summarize these findings through two key observations:

\paragraph{(a) Dimensional collapse.}  
As illustrated in Fig.~\ref{fig:motivation-b}, the singular value spectrum of the learned features under LogitNorm exhibits several small singular values approaching zero, indicating a collapse of feature variance along many dimensions. This suggests that LogitNorm may cause a potential loss of representational information, as features become concentrated in a few dominant directions.  
Such reduction in the effective feature dimensionality not only limits the expressiveness of the learned representation, but may also hinder downstream tasks that rely on rich feature diversity~\cite{hua2021feature}. In extreme cases, the network may overfit to a low-dimensional subspace, making it less robust to variations in the input and less capable of separating complex decision boundaries.

\paragraph{(b) Origin collapse.}  
We further observe that OOD data tend to reside closer to the origin in the feature space compared to ID data, implying that neural networks inherently produce lower-norm embeddings for OOD samples. To confirm this, we train a classifier with a feature dimensionality of $m=2$, \ie \( \mathbf{z} \in \mathbb{R}^2 \), to allow direct visualization. We prefer this approach over non-linear dimensionality reduction methods such as t-SNE~\cite{van2008visualizing} or UMAP~\cite{mcinnes2018umap}, which are highly sensitive to hyperparameters and often difficult to interpret. As shown in Fig.~\ref{fig:motivation-a}, when CIFAR-10 is used as the ID dataset and SVHN as the OOD dataset, the OOD samples are observed to collapse near the origin. We demonstrate how LogitNorm encourage this types of collapse in Section~\ref{subsec:logit_origin}.

\subsection{LogitNorm Further Encourages Collapse}
\label{subsec:logit_origin}
Building on the observations from the previous section, we aim to analyze the collapsing behavior of LogitNorm through its normalization factor, $\tau || \mathbf{f} ||$, in Eq.~(\ref{eqn:logitnorm}). To this end, we examine the phenomenon in the feature space, \( \mathbf{z} \in \mathbb{R}^m \), rather than directly normalizing the logits $\mathbf{f}$. This is justified by the proportional relationship between $||\mathbf{f}||$ and $||\mathbf{z}||$, indicating that normalization in logit space corresponds to an equivalent operation in the feature space.

\begin{proposition}
The norm of the logits \( ||\mathbf{f}|| \) is approximately proportional to the feature norm \( ||\mathbf{z}|| \), up to an additive noise term, such that \(||\mathbf{f}|| \approx \bar{\sigma} ||\mathbf{z}|| + \eta\), where $\bar{\sigma}$ is the weighted mean of the singular values.
More formally, the following bound holds:
\[
\sigma_{\min} ||\mathbf{z}|| - ||\mathbf{b}|| \leq ||\mathbf{f}|| \leq \sigma_{\max} ||\mathbf{z}|| + ||\mathbf{b}||,
\]
where \( \sigma_{\min} \) and \( \sigma_{\max} \) denote the smallest and largest singular values of the weight matrix \( \mathbf{W} \), respectively.
\label{prop:fz}
\end{proposition}

\noindent
The proof of Proposition~\ref{prop:fz} is provided in Appendix A. It shows that the LogitNorm objective can be approximated as,

\begin{equation}
\mathcal{L}_{\text{LogitNorm}}(f(\mathbf{x}; \theta), y) \approx - \log \frac{e^{f_y/\hat{\tau} || \mathbf{z} ||}}{\sum_{i=1}^{c} e^{f_i/\hat{\tau} || \mathbf{z} ||}},
\end{equation}
where \(\hat{\tau}\) is a new hyperparameter. We evaluate the proportionality between \( ||\mathbf{f}|| \) and \( ||\mathbf{z}|| \) through experiments on ImageNet-1K, ImageNet-200, CIFAR-100, and CIFAR-10 using ResNet18 and ResNet50. Empirical results show that the norm \( ||\mathbf{b}|| \) lies within the range \([0.05, 0.35]\), and the relative magnitude \( \frac{||\mathbf{b}||}{\sigma_{\max}} \) remains around \(2\%\)–\(4\%\). This indicates that the additive term \( ||\mathbf{b}|| \) introduces only a marginal shift in \( ||\mathbf{f}|| \), with the observed proportionality primarily governed by singular values, confirming the tightness of the bound. As shown in Fig.~\ref{fig:motivation-c}, there exists a proportional relationship between \( ||\mathbf{f}|| \) and \( ||\mathbf{z}|| \).  \( ||\mathbf{z}|| \) represents the distance to the origin, which implies that LogitNorm implicitly enforces the network based on feature distance from the origin, thereby encouraging feature collapse toward the origin.

% \begin{figure}[t]
%   \centering
%   %\fbox{\rule{0pt}{2in} \rule{0.9\linewidth}{0pt}}
%   \includegraphics[width=1\linewidth]{ICCV2025-Author-Kit-Feb/figs/CIFAR100_ce.png}
%    \caption{\(||\mathbf{f}||\) v.s. \(||\mathbf{z}||\)}
%    \label{fig:prop}
% \end{figure}
%

%\subsection{Method}

\subsection{Extended Logit Normalization}
We have  empirically and theoretically shown that LogitNorm implicitly constrains  feature representations based on their distance from the origin \( ||\mathbf{z}|| \), which can lead to feature collapse.   Alternatively, this notion of distance can be generalized from the origin to the decision boundaries. Intuitively, samples closer to a boundary exhibit higher uncertainty, while those farther away are classified with greater confidence. Therefore, a well-calibrated feature representation should capture each sample’s proximity to the decision boundaries rather than collapsing toward the origin.

\noindent To this end, we replace the scaling factor based on the distance to a singular point (the origin) with one based on the distance to the decision boundaries, introducing a new scaling mechanism. This extends LogitNorm into a more general formulation, which we refer to as \textbf{ELogitNorm}. Let \( f_{\max} \) denote the index of the maximum entry in the logit vector \( \mathbf{W}^\top \mathbf{z} + \mathbf{b} \), defined as:
\begin{equation}
    f_{\max} = \arg\max_{i \in \{1, \dots, c\}} \left( \mathbf{W}^\top \mathbf{z} + \mathbf{b} \right)_i.
\end{equation}
Then, the average distance to decision boundaries for a given feature vector \( \mathbf{z} \), whose predicted class corresponds to \( f_{\max} \), is defined as a point to a plane equation:
\begin{equation}
    \mathcal{D}(\mathbf{z}) := \frac{1}{c-1} \sum_{i \neq f_{\max}}^{c} 
    \frac{\left| (\mathbf{w}_{f_{\max}} - \mathbf{w}_i)^T \mathbf{z} + (b_{f_{\max}} - b_i) \right|}
    {\|\mathbf{w}_{f_{\max}} - \mathbf{w}_i\|_2},
\end{equation}
where \( f_{\max} \) is the predicted class index, and the summation excludes the term corresponding to \( f_{\max} \). We replace the scaling factor in Eq.~(\ref{eqn:logitnorm}) with \( s = \mathcal{D}(\mathbf{z}) \), leading to our \textbf{training objective}:
\begin{equation}
    \mathcal{L}_{\text{ELogitNorm}}(f(\mathbf{x}; \theta), y) = - \log \frac{e^{f_y/ \mathcal{D}(\mathbf{z})}}{\sum_{i=1}^{k} e^{f_i/ \mathcal{D}(\mathbf{z})}}.
    \label{eqn:elogitnorm}
\end{equation}
LogitNorm implicitly uses \(||\mathbf{z}||\) as a proxy for margin, we replace it with the actual multi-class margin, computed via average point-to-plane distances to all competing classes.
\noindent
\begin{proposition}
(Dimension of Minimum Scaling Factor Space) If \( m \geq c -1 \), the minimum distance to decision boundaries \(\mathcal{D}_{\min}(\mathbf{z}) = 0\) is attained when \(\mathbf{z} \in \bigcap_{i \neq f_{\max}} H_{i~f_{\max}}\), where \( H_{i~f_{\max}} \) denotes the decision boundary between the predicted class \( f_{\max} \) and class \( i \). This intersection forms an affine subspace of dimension \( m - c + 1 \). 
\label{prop:space}
\end{proposition}

\noindent
From Proposition~\ref{prop:space} (see proof in Appendix~B), the minimum scaling factor space in \textbf{ELogitNorm} exhibits a substantially higher dimensionality compared to that of LogitNorm. For instance, in a ResNet-18 model trained on CIFAR-10, where the feature dimension is \( m = 512 \) and the number of classes is \( c = 10 \), the resulting affine subspace has a dimension of \( 503 \). In contrast, LogitNorm enforces a minimum norm constraint \( ||\mathbf{z}||_{\min} = 0 \), which corresponds to a singular point at the origin.  Since the minimum scaling factor space represents the optimal region for neural network optimization, our method effectively prevents collapse to a singular point. As illustrated in Fig.~\ref{fig:motivation-b}, the singular value spectrum in \textbf{ELogitNorm} is more evenly distributed, avoiding the dominance of a few singular vectors. Furthermore, as demonstrated in Section~\ref{sec:experiments} and Fig.~\ref{fig:ex}, our proposed hyperparameter-free \textbf{ELogitNorm} consistently improves multiple post-hoc OOD scoring methods, whereas LogitNorm often underperforms or fails to generalize effectively.

\section{Experiments}
\label{sec:experiments}
In this section, we describe the experimental setup in detail and evaluate our method on several standard benchmarks, including small-scale OOD benchmarks (\ie. CIFAR-10 and CIFAR-100), and a large-scale OOD benchmark (\ie. ImageNet-200 and ImageNet-1k). We closely follow the OpenOOD evaluation\footnote{The codebase is: \href{https://github.com/Jingkang50/OpenOOD}{https://github.com/Jingkang50/OpenOOD}}~\cite{zhang2023openood}. All experiments are repeated three times with different random seeds, and we report the average performance. We run our experiments on an NVIDIA A100.
%and implement all methods using OpenOOD in PyTorch.

\begin{table*}[t]
\centering
\caption{\emph{Per-Dataset Performance of OOD Detection Methods and Their ELogitNorm-Enhanced Variants (denoted with $*$).} 
The image encoder is ResNet18. The ID dataset is \textbf{CIFAR-10}. \textcolor{myblue}{Blue} indicates \textcolor{myblue}{improvement}, and \textcolor{myorange}{orange} indicates \textcolor{myorange}{degradation}. }  
\begin{adjustbox}{width=\linewidth,center}
\begin{tabular}{llccccllccccccccll}
\toprule
& \multirow{2}{*}{Method}   &  
\multicolumn{2}{c}{\textbf{CIFAR-100}} &
\multicolumn{2}{c}{\textbf{TIN}} &
\multicolumn{2}{c}{\textbf{Near-OOD}} &
\multicolumn{2}{c}{\textbf{MNIST}} & 
\multicolumn{2}{c}{\textbf{SVHN}} &
\multicolumn{2}{c}{\textbf{Textures}} &
\multicolumn{2}{c}{\textbf{Places365}} &
\multicolumn{2}{c}{\textbf{Far-OOD}} \\
& & AUROC & FPR & AUROC & FPR & AUROC & FPR & AUROC & FPR & AUROC &  FPR & AUROC &  FPR & AUROC &  FPR & AUROC &  FPR \\
\doublerule

& \MSP &87.19 & 53.08 & 88.87 & 43.27 & 88.03 & 48.17 & 92.63 & 23.64 & 91.46 & 25.82 & 89.89 & 34.96 & 88.92 & 42.47 & 90.73 & 31.72 \\
\rowcolor{mygray} & \MSPTAG & 91.05 & 33.27 & 93.77 & 23.99 & 92.89 \textcolor{myblue}{\sr{4.86}} & 26.49\textcolor{myblue}{\sg{21.68}} & 99.12 & 3.96 & 98.31 & 7.66 & 94.61 & 23.21 & 94.22 & 22.81 & 96.68 \textcolor{myblue}{\sr{5.95}} & 13.73 \textcolor{myblue}{\sg{17.99}} \\
\midrule
& \ReAct & 85.93 & 67.40 & 88.29 & 59.71 & 87.11 & 63.56 & 92.81 & 33.77 & 89.12 & 50.23 & 89.38 & 51.42 & 90.35 & 44.20 & 90.42 & 44.90 \\
\rowcolor{mygray} & \ReActTAG & 90.79 & 33.97 & 93.83 & 23.74 & 92.31 \textcolor{myblue}{\sr{5.20}} & 28.86 \textcolor{myblue}{\sg{34.70}} & 99.44 & 2.53 & 98.45 & 7.13 & 94.79 & 22.77 & 94.43 & 22.54 & 96.78 \textcolor{myblue}{\sr{6.36}} & 13.74 \textcolor{myblue}{\sg{31.16}} \\
\midrule
& \KNN & 89.73 & 37.64 & 91.56 & 30.37 & 90.64 & 34.01 & 94.26 & 20.05 & 92.67 & 22.60 & 93.16 & 24.06 & 91.77 & 30.38 & 92.96 & 24.27 \\
\rowcolor{mygray} & \KNNTAG & 90.96 & 33.23 & 93.29 & 25.88 & 92.56 \textcolor{myblue}{\sr{1.92}} & 28.06 \textcolor{myblue}{\sg{5.95}} & 98.53 & 7.20 & 97.81 & 11.44 & 95.74 & 19.82 & 93.96 & 24.07 & 96.60 \textcolor{myblue}{\sr{3.64}} & 15.19 \textcolor{myblue}{\sg{9.08}} \\
\midrule
& \GEN &87.21 & 58.75 & 89.20 & 48.59 & 88.20 & 53.67 & 93.83 & 23.00 & 91.97 & 28.14 & 90.14 & 40.74 & 89.46 & 47.03 & 91.35 & 34.73 \\
\rowcolor{mygray} & \GENTAG & 90.96 & 33.58 & 93.88 & 23.83 & 92.42 \textcolor{myblue}{\sr{4.22}} & 28.70 \textcolor{myblue}{\sg{24.97}} & 99.41 & 2.74 & 98.41 & 7.29 & 94.67 & 23.48 & 94.44 & 22.50 & 96.73 \textcolor{myblue}{\sr{5.38}} & 14.00 \textcolor{myblue}{\sg{20.73}} \\
\midrule
& \fDBD & 89.56 & 39.61 & 91.65 & 30.57 & 90.61 & 35.09 & 94.71 & 19.33 & 92.93 & 22.50 & 93.13 & 24.35 & 92.01 & 29.15 & 93.19 & 23.84 \\
\rowcolor{mygray} & \fDBDTAG & 90.31 & 36.28 & 93.44 & 24.72 & 91.87 \textcolor{myblue}{\sr{1.26}} & 30.50 \textcolor{myblue}{\sg{4.59}} & 99.28 & 3.31 & 98.67 & 6.05 & 95.78 & 17.93 & 93.94 & 23.23 & 96.92 \textcolor{myblue}{\sr{3.73}} & 12.63 \textcolor{myblue}{\sg{11.21}} \\
\midrule
& \SCALE & 81.27 & 81.78 & 83.98 & 78.87 & 82.62 & 80.32 & 90.58 & 48.69 & 84.91 & 70.17 & 83.93 & 80.54 & 86.41 & 70.57 & 86.46 & 67.49 \\
\rowcolor{mygray} & \SCALETAG & 90.86 & 34.17 & 93.86 & 23.75 & 92.36 \textcolor{myblue}{\sr{9.74}} & 28.96 \textcolor{myblue}{\sg{51.36}} & 99.54 & 2.09 & 98.78 & 5.79 & 95.23 & 21.63 & 94.19 & 23.22 & 96.94 \textcolor{myblue}{\sr{10.48}} & 13.18 \textcolor{myblue}{\sg{54.31}} \\
\bottomrule
\end{tabular}
\end{adjustbox}
 
\label{tab:CIFAR10-main}
\end{table*}

\begin{table*}[h]
\centering 
\begin{adjustbox}{width=\linewidth,center}
\begin{tabular}{llcccccccccccccc}
\toprule
& \multirow{2}{*}{Method}   &  
\multicolumn{2}{c}{\textbf{SSB-hard}} &
\multicolumn{2}{c}{\textbf{NINCO}} &
\multicolumn{2}{c}{\textbf{Near-OOD}} &
\multicolumn{2}{c}{\textbf{iNaturalist}} & 
\multicolumn{2}{c}{\textbf{Textures}} &
\multicolumn{2}{c}{\textbf{OpenImage-O}} &
\multicolumn{2}{c}{\textbf{Far-OOD}} \\
& & AUROC & FPR & AUROC & FPR & AUROC & FPR & AUROC & FPR & AUROC &  FPR & AUROC &  FPR & AUROC &  FPR  \\
\doublerule

& Cross-Entropy & \textbf{72.09} & \textbf{74.49} & 79.95 & 56.88 & 76.02 & 65.68 & 88.41 & 43.34 & 82.43 & 60.87 & 84.86 & 50.13 & 85.23 & 51.45 \\

& LogitNorm~\cite{logitnorm} & 67.50 & 82.08 & 81.73 & 55.04 & 74.62 & 68.56 & 94.57 & 20.75 & 89.30 & 40.82 & 90.75 & 32.38 & 91.54 & 31.32 \\

\rowcolor{mygray} & \textbf{ELogitNorm}& 70.52 & 76.85 & \textbf{83.24} & \textbf{52.22} & \textbf{76.88} & \textbf{64.54} & \textbf{96.15} & \textbf{16.56} & \textbf{91.46} & \textbf{36.58} & \textbf{91.97} & \textbf{30.07} & \textbf{93.19} & \textbf{27.74} \\
\bottomrule
\end{tabular}
\end{adjustbox}
\caption{\emph{Per-Dataset Performance of OOD Detection Training Methods including Cross-Entropy, LogitNorm~\cite{logitnorm}, and \textbf{ELogitNorm} (Ours).} 
The image encoders is ResNet50. The ID dataset is \textbf{ImageNet1K}. \MSP is used for OOD score.} 
\label{tab:ImageNet1k-main}
\end{table*}

\subsection{Experiment settings}

\paragraph{Benchmark.}  
We evaluate \textit{ELogitNorm} using OpenOOD \cite{zhang2023openood}, which includes four vision benchmarks: \textit{CIFAR-10} \cite{cifar}, \textit{CIFAR-100} \cite{cifar}, \textit{ImageNet-200} \cite{Image-net-1k}, and \textit{ImageNet-1k} \cite{Image-net-1k} for training-based OOD detection methods.  Each benchmark consists of an in-distribution dataset \( \id \) and multiple out-of-distribution (OOD) datasets \( \ood \), further categorized into \textit{Near-OOD} and \textit{Far-OOD} datasets. The distinction between "near" and "far" is based on the similarity of OOD samples to ID samples, with Near-OOD samples posing a greater challenge for separation. OpenOOD also provides pre-trained model checkpoints for CIFAR-10, CIFAR-100, ImageNet-200, and ImageNet-1k trained using standard Cross-Entropy loss. We primarily evaluate the improvements achieved by our training method over Cross-Entropy, while also comparing against alternative training methods under the same fixed post-hoc detection techniques.

\paragraph{Metric.} We employ two conventional metrics to evaluate OOD detection performance. The first is a threshold-independent metric: Area Under the Receiver Operating Characteristic Curve (AUROC), where higher percentages indicate better performance. The second metric is the False Positive Rate at 95\% True Positive Rate (FPR95), where lower percentages reflect better performance.  

\paragraph{Implementation Details.}
We train ResNet-18 on CIFAR-10/100 for 100 epochs and on ImageNet-200 for 90 epochs using SGD with momentum~0.9, weight decay $5\times10^{-4}$, batch size~128, and an initial learning rate of~0.1 with standard scheduling. For ImageNet-1K, we finetune a ResNet-50 for 30 epochs with a learning rate of~0.001. Our method is hyperparameter-free and adds only minimal computational overhead, with an efficient implementation that scales well even for large numbers of classes~\(c=1000\) (see Appendix).

\subsection{Main results}
In this section, we demonstrate the effectiveness of our method from three perspectives: 
(a) \emph{OOD detection}, where we evaluate its ability to enhance post-hoc methods and compare it with other training-based approaches, (b) \emph{model calibration}, measured by the expected calibration error and (c) in-distribution data \emph{classification accuracy}.

\begin{figure*}[h]
  \centering
    \caption{\emph{Average far-OOD and near-OOD performance with 4 post-hoc methods \MSP, \ReAct, \GEN and \SCALE.} ResNet18 are trained by \textcolor{eblue}{Cross-Entropy}, \textcolor{mygreen}{LogitNorm}~\cite{logitnorm}, and \textcolor{mysalmon}{\textbf{ELogitNorm} (Ours)}, respectively. ID data are CIFAR-10 and CIFAR-100.}
  \begin{subfigure}{0.24\linewidth}
    \includegraphics[width=1\linewidth]{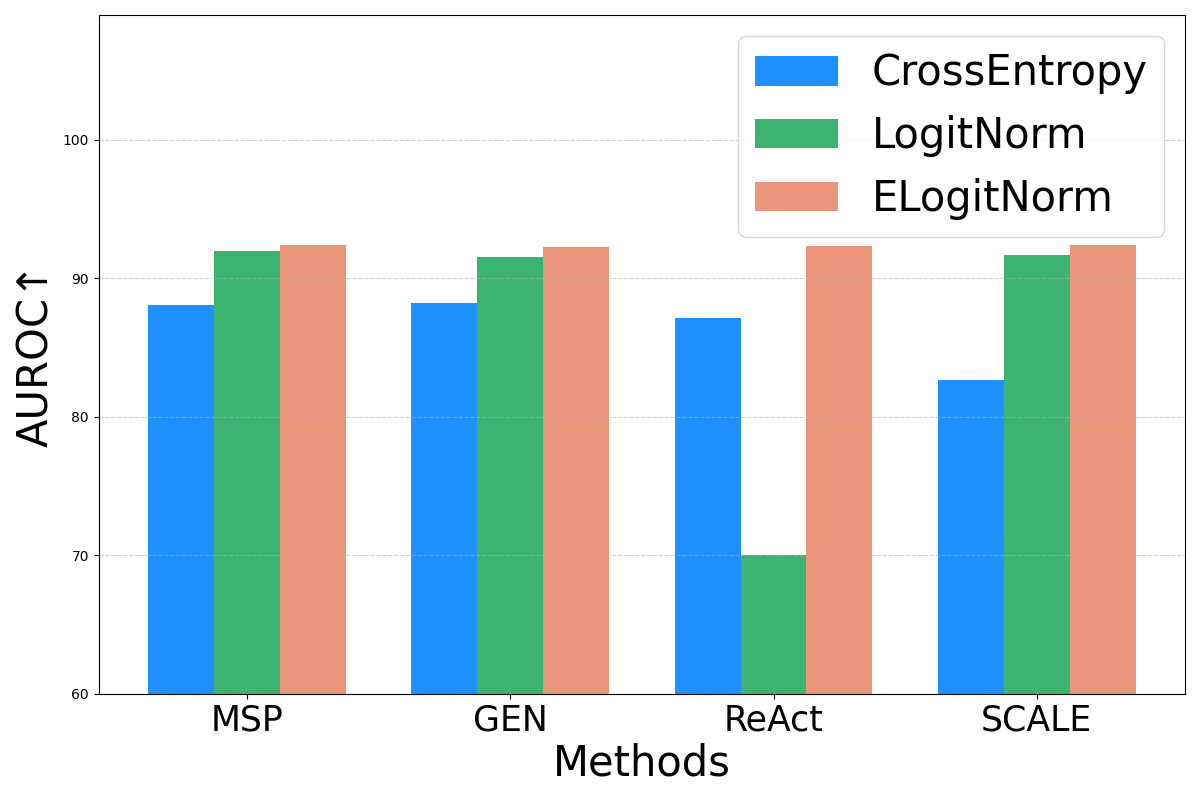}
    \caption{CIFAR-10, near-OOD}
    \label{fig:ex-a}
  \end{subfigure}
  \begin{subfigure}{0.24\linewidth}
    \includegraphics[width=1\linewidth]{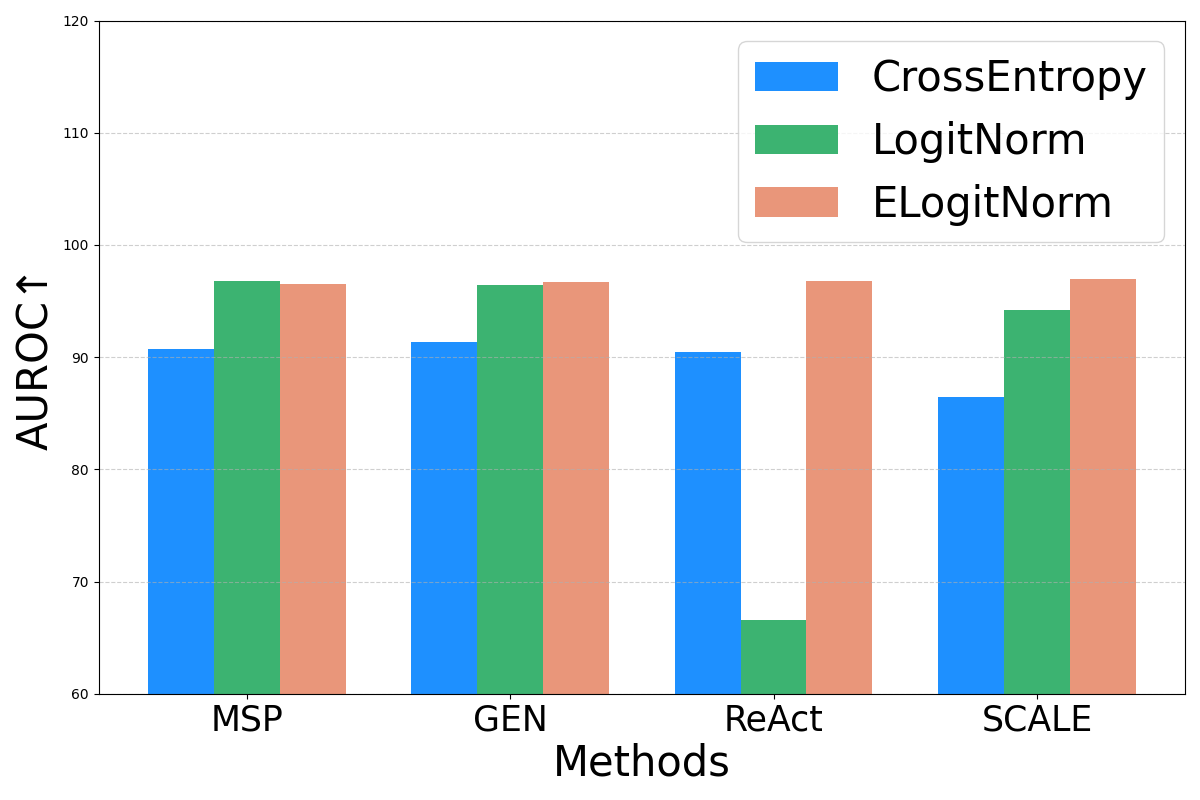}
    \caption{CIFAR-10, far-OOD}
    \label{fig:ex-b}
  \end{subfigure}
    \begin{subfigure}{0.24\linewidth}
    \includegraphics[width=1\linewidth]{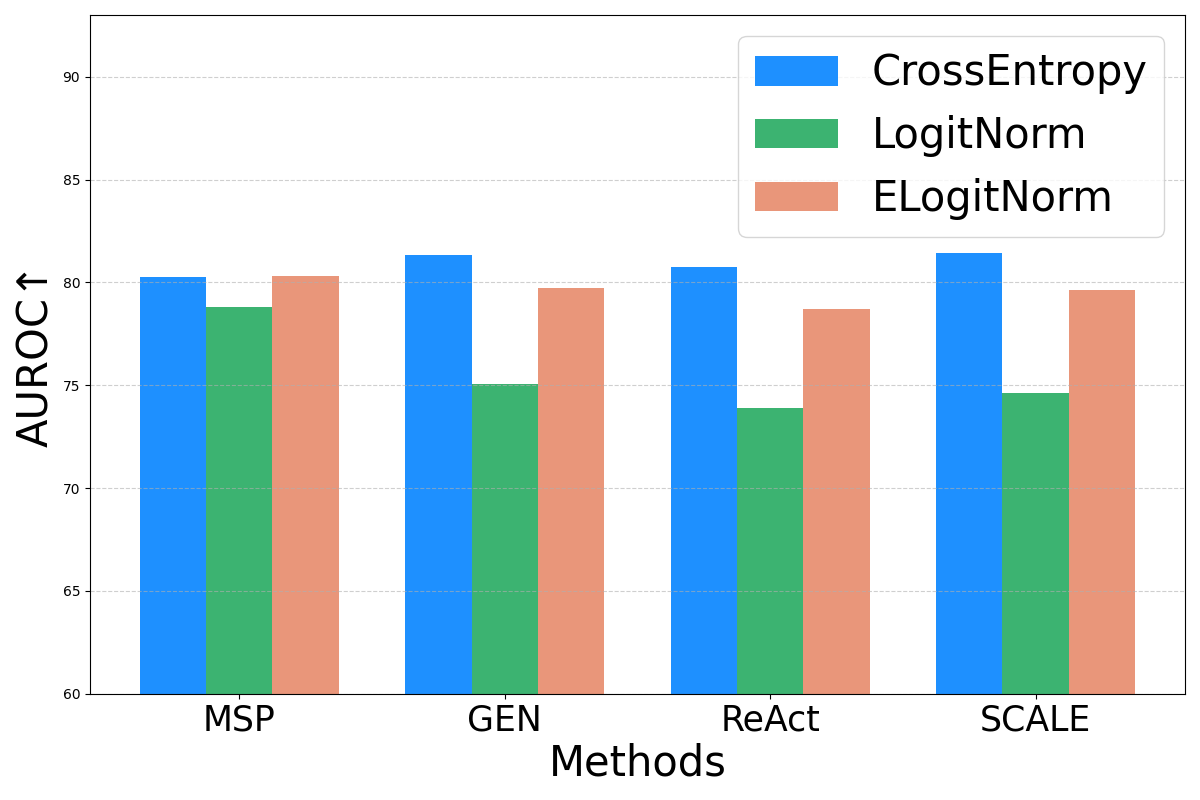}
    \caption{CIFAR-100, near-OOD}
    \label{fig:ex-c}
  \end{subfigure}
      \begin{subfigure}{0.24\linewidth}
    \includegraphics[width=1\linewidth]{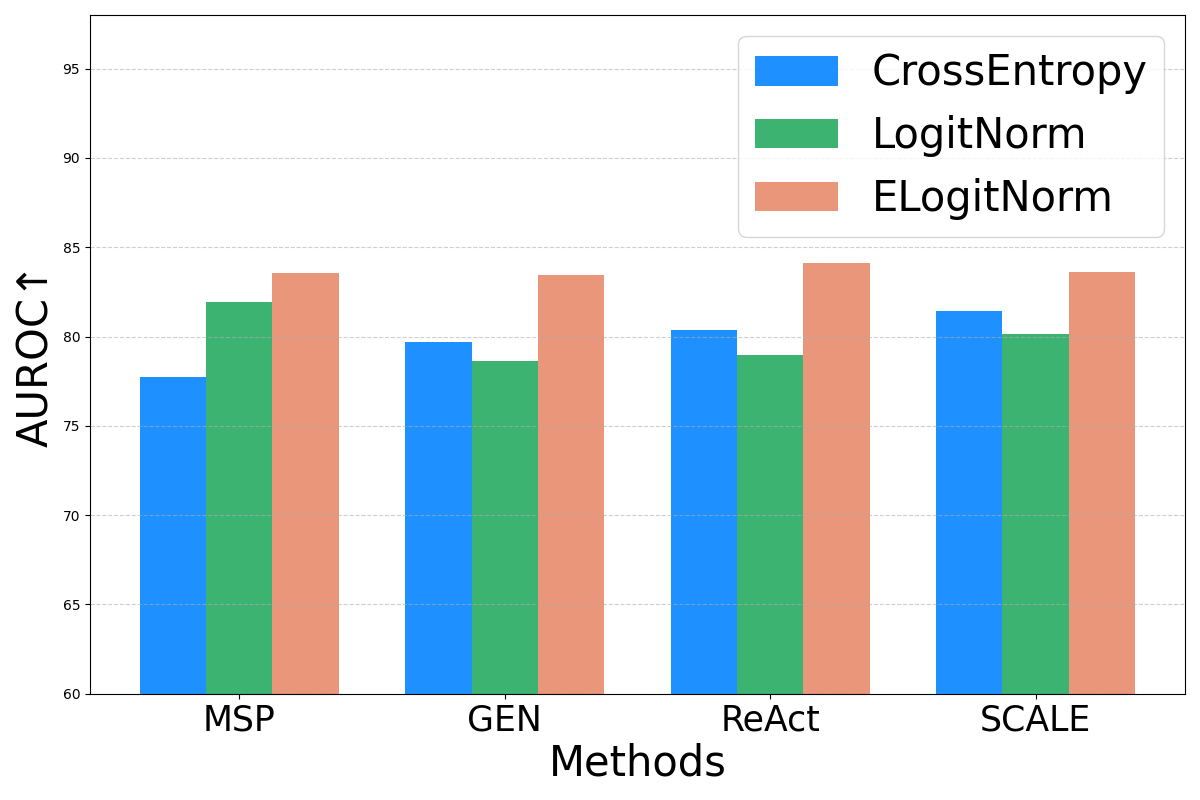}
    \caption{CIFAR-100, far-OOD}
    \label{fig:ex-d}
  \end{subfigure}

  \label{fig:ex}
\end{figure*}

\begin{table*}
\centering
\begin{adjustbox}{width=0.92\linewidth,center}
\begin{tabular}{llcccccccccccccc}
\toprule
& \multirow{4}{*}{Method}   &  
\multicolumn{4}{c}{\textbf{CIFAR10}} & 
\multicolumn{4}{c}{\textbf{CIFAR100}}&
\multicolumn{4}{c}{\textbf{ImageNet-200}}\\

& \multirow{2}{*}{Method}   &  
\multicolumn{2}{c}{\textbf{Near-OOD}} &
\multicolumn{2}{c}{\textbf{Far-OOD}} &
\multicolumn{2}{c}{\textbf{Near-OOD}} &
\multicolumn{2}{c}{\textbf{Far-OOD}} & 
\multicolumn{2}{c}{\textbf{Near-OOD}} &
\multicolumn{2}{c}{\textbf{Far-OOD}} \\
& & AUROC & FPR & AUROC & FPR & AUROC & FPR & AUROC & FPR & AUROC &  FPR & AUROC &  FPR  \\
\doublerule
& \textit{\MSP}\\
\midrule
& Cross-Entropy & 88.03 & 48.17 & 90.73 & 31.72 & \textbf{80.27} & \textbf{54.80} & 77.76 & 58.70 & \textbf{83.34} & \textbf{54.82} & 90.13 & 35.43 \\
& LogitNorm~\cite{logitnorm} & 92.33 & 29.34 & \textbf{96.74} & 13.81 & 78.47 & 62.89 & 81.53 & 53.61 & 82.66 & 56.46 & 93.04 & 26.11 \\
\rowcolor{mygray} & \textbf{ELogitNorm} (Our) & \textbf{92.89} & \textbf{26.49} & 96.68 & \textbf{13.73} & 79.68 & 59.48 & \textbf{84.51} & \textbf{46.86} & 83.06 & 54.99 & \textbf{93.58} & \textbf{25.08} \\
\doublerule
& \textit{\KNN}\\
\midrule
& Cross-Entropy & 90.64 & 34.01 & 92.96 & 24.27 & \textbf{80.18} & 61.22 & 82.40 & 53.65 & 81.57 & 60.18 & 93.16 & 27.27 \\
& \CIDER & 90.71 & 32.11 & 94.71 & 20.72 & 73.10 & 72.02 & 80.49 & 54.22 & 80.58 & 60.10 & 90.66 & 30.17 \\
& \NPOS & 89.78 & 32.64 & 94.07 & 20.59 & 78.35 & 63.35 & 82.29 & \textbf{51.13} & \textbf{84.37} & 62.09 & 94.83 & 21.76 \\
\rowcolor{mygray} & \textbf{ELogitNorm} (Our) & \textbf{92.56} & \textbf{28.06} & \textbf{96.60} & \textbf{15.19} & 80.12 & \textbf{58.10} & \textbf{82.84} & 54.23 & 82.73 & \textbf{56.65} & \textbf{96.08} & \textbf{18.04} \\
\bottomrule
\end{tabular}
\end{adjustbox}
\caption{\emph{Far-OOD and near-OOD performance from different training methods with fixed OOD scoring function.} 
The image encoders is ResNet18. The ID datasets are \textbf{CIFAR-10}, \textbf{CIFAR-100}, and \textbf{ImageNet-200}. \KNN and \MSP used as post-hoc methods.}  
\label{tab:compareall-main}
\end{table*}

\paragraph{ELogitNorm enhances post-hoc OOD detection.} The baseline post-hoc methods include OOD scoring techniques such as \MSP, \GEN, \KNN, \ReAct and \SCALE. Table~\ref{tab:CIFAR10-main} demonstrates that \textbf{ELogitNorm} significantly improves OOD detection performance in both AUROC and FPR95 across all methods. A notable observation is that \textbf{ELogitNorm} leads to substantial improvements in far-OOD detection compared to near-OOD detection. For instance, in \SCALE, the AUROC for far-OOD datasets improves by +10.48\%, while the FPR95 is reduced by 54.31\%. This pattern is consistent across other methods, where enhancements are more pronounced in far-OOD scenarios. Results for CIFAR-100 and ImageNet-200 show that the improvements are particularly noticeable in far-OOD datasets and \MSP shows minor performance trade-offs on some near-OOD datasets, details can be found in the supplementary material.

\paragraph{ELogitNorm enables better enhancement than LogitNorm and alternatives.}
As shown in Table~\ref{tab:ImageNet1k-main}, \textbf{ELogitNorm} consistently improves OOD detection on ImageNet-1K, with especially strong gains in far-OOD (FPR95 reduced from 51.45\% to 27.74\%), exceeding the improvement of LogitNorm. For near-OOD, \textbf{ELogitNorm} remains stable even when LogitNorm degrades. Similar trends appear on CIFAR-10/100, where both methods offer improvements, but \textbf{ELogitNorm} delivers stronger and more reliable performance, avoiding the severe degradation LogitNorm encounters when combined with \ReAct\ (Section~\ref{sec:method}). Results on ImageNet-200 further confirm this robustness. As summarized in Table~\ref{tab:ImageNet1k-main} and Fig.~\ref{fig:ex}, \textbf{ELogitNorm} provides broader compatibility and stronger overall enhancement across post-hoc methods. When comparing with other training-based approaches under fixed post-hoc scoring (\MSP, \KNN; Table~\ref{tab:compareall-main}), \textbf{ELogitNorm} again achieves superior results. Despite \CIDER\ and \NPOS\ being designed for \KNN, our method attains higher performance—for example, an AUROC of 96.08 on ImageNet-200, surpassing \CIDER\ (90.66) and \NPOS\ (94.83). Although minor drops on near-OOD are observed, such degradation is common for all training-time methods~\cite{zhang2023openood}. Notably, \textbf{ELogitNorm} maintains more stable near-OOD performance than alternatives, representing a modest but meaningful step toward narrowing this long-standing performance gap.

\begin{table}
\centering
\begin{adjustbox}{width=0.7\linewidth,center}
\begin{tabular}{lccc}
\toprule
Method & \(\mathbf{f}\)  & \(\frac{\mathbf{f}}{\tau||\mathbf{f}||}\) & \(\frac{\mathbf{f}}{\mathcal{D}(\mathbf{z})}\) \\
\midrule
Cross-Entropy & \textbf{3.3} & 4.8 & 23.3 \\
LogitNorm~\cite{logitnorm}  & 58.7  &  \textbf{4.1}  & 52.3\\
\rowcolor{mygray}\textbf{ELogitNorm} (Ours) & 26.7 &  4.7 & \textbf{1.8}\\
\bottomrule
\end{tabular}
\end{adjustbox}
\caption{\emph{Expected Calibration Error (ECE) (\%).} The binning size is set to be 15.  The model is ResNet18 trained on CIFAR-10.}
\label{tab:ece_scores}
\end{table}

\begin{figure*}[t]
    \centering
    % ---- First minipage: Figure ----
    \begin{minipage}{0.48\textwidth}
        \centering
        \includegraphics[width=1.0\linewidth]{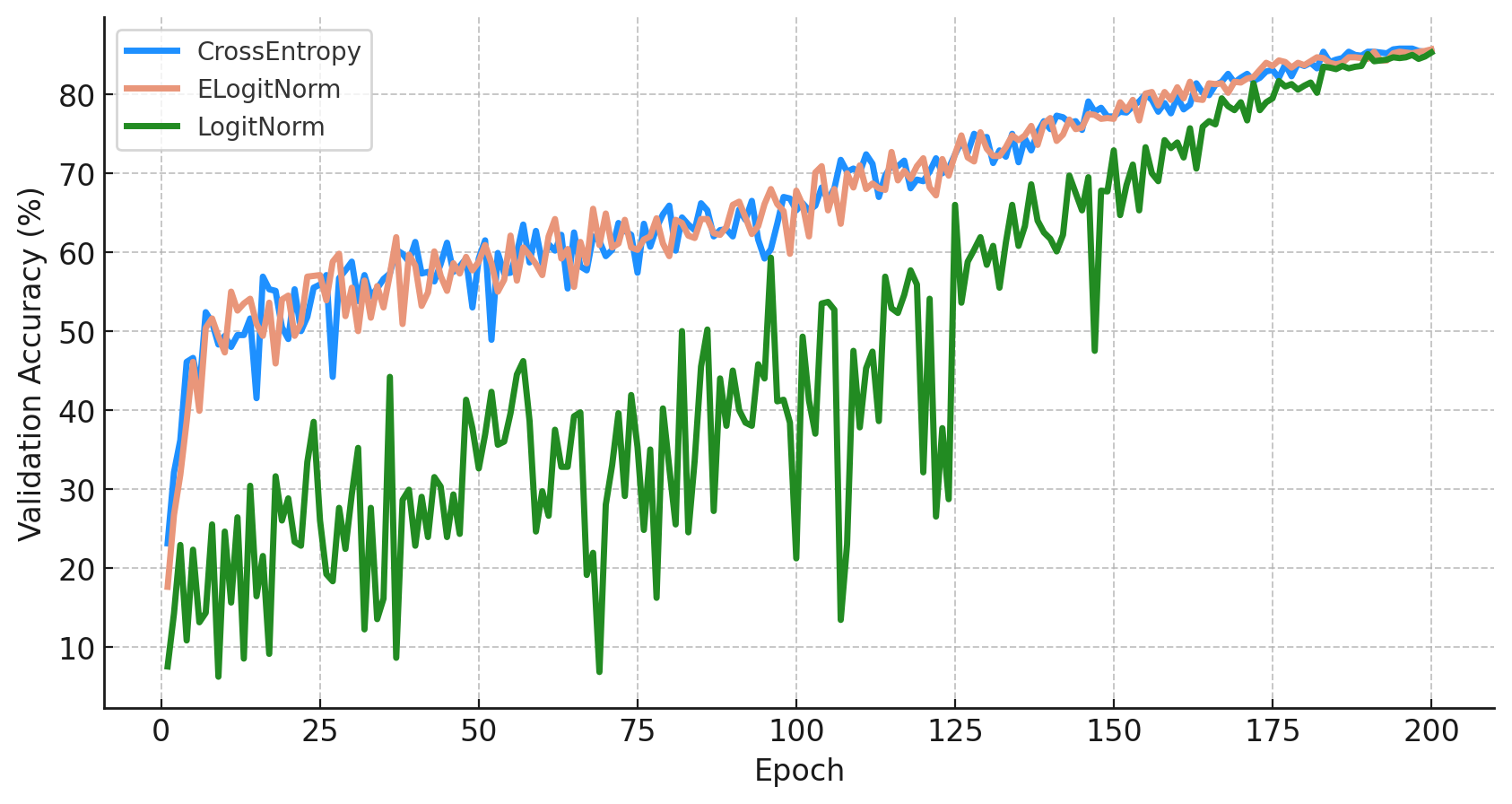}
        \captionof{figure}{\emph{Validation accuracy during training.} ResNet18 trained on ImageNet-200 over 200 training epochs. 
        The proposed \textcolor{mysalmon}{\textbf{ELogitNorm} (Ours)} achieves a stable training curve compared to \textcolor{mygreen}{LogitNorm}~\cite{logitnorm}, while maintaining competitive final accuracy as \textcolor{eblue}{Cross-Entropy}.}
        \label{fig:acc_cifar10}
    \end{minipage}
    \hfill
    % ---- Second minipage: Table ----
    \begin{minipage}{0.48\textwidth}
        \centering
        \begin{adjustbox}{width=\linewidth,center}
        \begin{tabular}{lccc}
            \toprule
            Method & CIFAR-10 & CIFAR-100 & ImageNet200 \\
            \midrule
            \multicolumn{4}{l}{\textit{100 Epochs / 90 Epoch for ImageNet200}} \\
            \midrule
            Cross-Entropy & \textbf{95.06} & \textbf{77.25} & 86.37\\
            LogitNorm~\cite{logitnorm}  & 94.30 & 76.34 & 86.04 \\
            \rowcolor{mygray}\textbf{ELogitNorm} (Ours) & 94.47 & 76.67 & \textbf{86.39} \\
            \midrule
            \multicolumn{4}{l}{\textit{200 Epochs}} \\
            \midrule
            Cross-Entropy & 95.10 & \textbf{77.47} & 86.58\\
            LogitNorm~\cite{logitnorm}  & 94.83 & 76.06 & 86.41 \\
            \rowcolor{mygray}\textbf{ELogitNorm} (Ours) & \textbf{95.11} & 77.37 & \textbf{87.12} \\
            \bottomrule
        \end{tabular}
        \end{adjustbox}
        \captionof{table}{\emph{ID Accuracy.} Comparison of Cross-Entropy, LogitNorm~\cite{logitnorm} and \textbf{ELogitNorm} (Ours) on \textbf{CIFAR-10}, \textbf{CIFAR-100} and \textbf{ImageNet-200} datasets.}
        \label{tab:elogitnorm_results}
    \end{minipage}
\end{figure*}

\paragraph{Calibration performance.} Expected calibration error (ECE) is a standard and widely adopted metric for quantifying confidence calibration in multi-class classifiers. Table~\ref{tab:ece_scores} reports ECE under various training objectives and logit-scaling strategies, allowing a controlled and comprehensive comparison. We consider three forms of scaled logits: the raw prediction vector $\mathbf{f}$, the normalized logits $\frac{\mathbf{f}}{\tau\lVert \mathbf{f}\rVert}$ derived from LogitNorm with a learned temperature~$\tau$, and the boundary-aware scaling $\frac{\mathbf{f}}{\mathcal{D}(\mathbf{z})}$, where $\mathcal{D}(\mathbf{z})$ measures the average distance from input $\mathbf{z}$ to nearby decision boundaries. This setup highlights how calibration quality depends jointly on the training loss and the normalization applied at inference. For instance, cross-entropy achieves its best calibration performance when evaluated with unscaled logits, while LogitNorm attains its lowest ECE when logits are normalized by the product of the learned temperature and their magnitude. Across all settings, however, our method demonstrates consistently superior calibration, achieving the lowest ECE regardless of the loss function or scaling strategy, thereby indicating improved robustness and stability in calibration performance.

\paragraph{Classification accuracy.}
A reliable classifier should not only detect OOD samples but also maintain strong ID performance. Table~\ref{tab:elogitnorm_results} reports accuracy on CIFAR-10, CIFAR-100, and ImageNet-200. The model trained with LogitNorm shows a clear degradation in accuracy compared to the standard cross-entropy baseline across all datasets. In contrast, \textbf{ELogitNorm} attains the highest accuracy on CIFAR-10 and ImageNet-200. As illustrated in Fig.~\ref{fig:acc_cifar10}, LogitNorm also displays unstable training behavior and inferior classification performance relative to the other two methods. Additional results for the remaining datasets are provided in the Appendix.

\section{Discussion}
\label{sec:discussion}

\paragraph{Distance Awareness.}  
Distance awareness has been extensively studied in OOD detection. \KNN estimates OOD likelihood based on the distance to the nearest neighbor, while \fDBD, closely related to our approach, assumes that OOD samples lie near decision boundaries and designs its scoring function accordingly. \CIDER also incorporates distance awareness by optimizing feature representations to maximize inter-class separation and enforce intra-class compactness. Inspired by these approaches, we extend the formulation of LogitNorm by incorporating distances to decision boundaries, leading to our proposed method, \textbf{ELogitNorm}. By amplifying distance awareness during training, \textbf{ELogitNorm} significantly enhances distance-based OOD scoring techniques such as \KNN and \fDBD. This perspective opens a broader direction for refining feature representations through alternative distance metrics to further improve OOD detection. By examining the relationship between ID and OOD data in the classifier’s representation space, we advocate for a deeper understanding of representation geometry and distance-based separation within learned feature spaces.

\paragraph{Adaptive temperature scaling}  
Temperature scaling~\cite{guo2017calibration} applies a fixed scaling factor to all logits, whereas both LogitNorm and ELogitNorm adopt an \textit{adaptive temperature scaling} mechanism, where the scaling factor varies for each sample. We define this adaptive scaling in a general form:
\begin{equation}
\mathcal{L}_{s}(f(\mathbf{x}; \theta), y) = - \log \frac{e^{f_y / s}}{\sum_{i=1}^{c} e^{f_i / s}},
\label{eqn:snorm}
\end{equation}
where \( s = \tau || \mathbf{f} || \) in Eq.~(\ref{eqn:logitnorm}) and \( s = \mathcal{D}(\mathbf{z}) \) in Eq.~(\ref{eqn:elogitnorm}). The training of both objectives can thus be interpreted as a form of dynamic calibration governed by the sample-dependent scaling factor \( s \). Notably, in LogitNorm, \( s = \tau || \mathbf{f} || \) depends solely on the logits \( \mathbf{f} \), whereas in ELogitNorm, \( s = \mathcal{D}(\mathbf{z}) \) incorporates information from the feature space. This allows ELogitNorm to achieve better calibration by explicitly accounting for each sample’s relative position within the feature space. By aligning feature scaling with decision boundaries, ELogitNorm naturally enhances OOD separation while maintaining strong discriminative power for ID classification. This perspective naturally connects to uncertainty estimation, where confidence scores should ideally reflect a sample’s epistemic uncertainty. \textbf{ELogitNorm} scales logits with attention on decision boundaries, making the model more sensitive to distribution shifts. Given the definition of \textit{adaptive temperature scaling} \( s \), one can design alternative scaling strategies to further enhance OOD detection and confidence calibration, potentially integrating concepts from uncertainty-aware representations \cite{mucsanyi2024benchmarking}.

\section{Related work}
\label{sec:relatedwork}

\paragraph{OOD Detection} OOD detection has been extensively studied form the perspective of 1) \emph{ designing OOD scoring functions} with access to information available at different stages of a pre-trained classifier. Maximum softmax probability (MSP)~\cite{DanMSP} and GEN~\cite{xixi2023GEN} operate with the information in the probability space (\ie. after $\softmax$ layer). Max-Logit~\cite{KL-matching}, energy score~\cite{energybased_ood} and NN-guide~\cite{nn-guide} utilize information form logit.
%Max-Logit~\cite{KL-matching} argues that the logits contain more information compared to the probability space. The energy score~\cite{energybased_ood} reinterprets the logits as the density of the unnormalized conditions, and the resulting score can be regarded as the density of the unnormalized data. 
%Formally, it reads as $\logsumexp$ all the logits (\ie, marginalizing over the label space). 
%GEN~\cite{xixi2023GEN} instead utilizes the distance between two distributions (\ie, predictive distribution and uniform distribution) as the OOD score. 
%The motivation is that OOD samples tend to generate a uniform distribution. 
Other OOD scoring methods operate in the feature space and commonly require access to the statics of the training data~\cite{kamoi2020mahalanobis, KL-matching, liu2023fast, neco, ahmadian2024unsupervised, ding2025revisiting}, which restricts their applicability. ViM~\cite{ViM} combines information from both feature and logit space. 
%fDBD~\cite{liu2023fast} assumes that the features of ID samples should be far away from the decision boundary among each class. Therefore, the final OOD score read as the average distance between the input among the per-class decision boundary. 
%KNN~\cite{sun2022knnood} is  computationally heavier compared to Mahalanobis distance, but it does not assume a Gaussian distribution for the per-class ID features. 
2) \emph{reshaping the extracted features} to enhance the OOD performance. Such methods typically work well in a post-hoc manner, meaning that they do not interfere with the training procedure~\cite{ODIN, ReAct, xu2024scaling, ash, rank}. Particularly,
%For example, ODIN~\cite{ODIN} proposes to modify the inputs by adding a perturbation inspired by the adversarial samples. 
ReAct~\cite{ReAct} observes that the OOD features tend to behave irregularly large in the feature space (\ie, the features extracted before the penultimate layer). To rectify the activations in a ``normal'' range, ReAct~\cite{ReAct} proposes to clip the features if their magnitude is above the $p$-th percentile of activations estimated on the ID data. 
%RankFeat~\cite{rank} observes that OOD features extracted from the intermediate layers tend to have a dominant singular value compared to the ID features. Therefore, it proposes to remove the rank-1 feature before calculating the energy score~\cite{energybased_ood}. 
ASH~\cite{ash} and SCALE~\cite{xu2024scaling} apply similar rescaling strategies to the features extracted from the penultimate layer, and they both need to access the statistics of the training data to set the rescaling factor. The difference is that ASH~\cite{ash} only rescales a specified portion of the features, whereas SCALE~\cite{xu2024scaling} does not. SCALE can also be incorporated at the training stage. However, it exhibits limited performance on small-scale OOD benchmark, cf. Fig.~\ref{fig:onecol}. Our method not only improves the performance of baseline OOD scoring methods, but also further boosts the performance of enhancing methods. 
3) \emph{training-loss modification}. Such methods either add a regularization loss in addition to cross entropy loss~\cite{outlier_exposure_for_DNN, du2023dream, du2022vos, csi} or reformulate the problem to be a deep-metric learning problem~\cite{cider, npos}. It is worth to note that he regularization loss commonly accounts for the real OOD samples~\cite{outlier_exposure_for_DNN} or synthesized OOD samples/features~\cite{du2022vos, du2023dream}. However, it commonly involves two-stage training (\ie, one stage for training a good multi-class classifier and generating OOD features/samples, one stage for training a binary classifier for differentiating ID and OOD samples.) which makes it less favorable in practice. Meanwhile,~\cite{pereyra2017regularizing} and ~\cite{logitnorm} achieve end-to-end training, resulting in an improved OOD detection performance and a better-calibrated classifier. 
Our method falls into this category, and its training is completely hyperparameter-free.
Compared to LogitNorm~\cite{logitnorm}, our method is free from a held-out dataset for selecting a proper value of temperature.  Meanwhile, our method works well with a wide range of OOD scoring including MSP~\cite{DanMSP}, KNN~\cite{sun2022knnood}, GEN~\cite{xixi2023GEN}, and fDBD~\cite{liu2023fast} as well as enhancing methods such as ReAct~\cite{ReAct} and SCALE~\cite{xu2024scaling}. Feature normalization has been explored as a post-hoc strategy in~\cite{park2023understanding,yu2023block}. 
T2FNorm~\cite{t2fnorm} operates in both training and post-hoc stages, yet it does not demonstrate compatibility with other post-hoc methods and lacks confidence calibration. 
Beyond these approaches, several works jointly address OOD generalization and detection including Scone~\cite{bai2023feed}, AHA~\cite{bai2024aha} and InfoBound~\cite{zhu2025infobound}.

\paragraph{Calibration for Classification} Modern neural networks trained with trained with cross-entropy loss are prone to be overconfident~\cite{calibration_modern_nn}. Later work~\cite{xiong2023proximityinformed} analyzes predictions per sample and empirically shows that models tend to be under-confident for lower proximity samples and over-confident for higher proximity samples. To mitigate the issue of miscalibration, there are roughly two types of methods consisting of 1) post hoc methods and 2) training loss modification. Post hoc methods can be divided into scaling-based methods that include temperature scaling, parameterized temperature scaling~\cite{PTS}, and Mix-n-Match~\cite{mixnmatch} and binning-based methods such as classic histogram binning~\cite{histogram_binning}, mutual information maximization-based binning~\cite{MI_binning}, and isotonic regression~\cite{isotonic_regression}. Another line of work requires training with additional loss such as ~\cite{pereyra2017regularizing} and logit-norm~\cite{logitnorm}.

\paragraph{Feature collapse.}
Feature collapse occurs when learned representations degenerate into a low-diversity, low-rank subspace that fails to distinguish samples from different classes. In self-supervised learning, this behavior has been illustrated by \cite{hua2021feature}, who showed that whitening batch normalization can mitigate collapse. A theoretical analysis of dimensional collapse in contrastive methods was provided by \cite{jing2021understanding}, and several subsequent works connected this phenomenon to performance degradation \cite{he2022exploring, ghosh2022investigating, li2022understanding, garrido2023rankme}. Related issues also arise in generative modeling: GANs are well known to exhibit mode collapse~\cite{arjovsky2017wasserstein, miyato2018spectral}, and vector-quantized models can similarly suffer from feature collapse that reduces generative quality~\cite{zhu2025addressing}. Overall, preventing collapse requires balancing representation compactness with sufficient expressiveness so that embeddings retain meaningful semantic structure rather than converging to trivial solutions. This balance is also critical for OOD detection, where overly uniform or over-regularized features reduce inter-class separation and degrade sensitivity.

\section{Conclusion}
\label{sec:conclusion}
In this paper, we introduced \textbf{ELogitNorm}, a generalized and hyperparameter-free extension of LogitNorm that leverages distances to decision boundaries in the feature space of neural network classifiers. Our approach effectively mitigates the feature collapse issue inherent in LogitNorm, leading to improved OOD detection, better confidence calibration, and preserved classification accuracy.  Extensive experiments demonstrate that \textbf{ELogitNorm} consistently enhances the OOD detection capability of post-hoc methods and achieves superior performance compared to prior training-based approaches, particularly on far-OOD benchmarks. While the improvements on near-OOD settings are more modest, the results remain robust and stable across architectures and datasets.  Beyond performance gains, this work underscores the importance of understanding the geometric structure of feature representations in deep networks. We hope that \textbf{ELogitNorm} will serve as a foundation for future research exploring boundary-aware calibration, adaptive scaling mechanisms, and more principled ways to align feature geometry for reliable open-world recognition.

\section*{Acknowledgement}
This work was partially supported by the Wallenberg AI, Autonomous Systems and Software Program (WASP) funded by the Knut and Alice Wallenberg Foundation, and the Zenith career development program at Linköping University. The computational resources were provided by the National Academic Infrastructure for Supercomputing in Sweden (NAISS) at the National Supercomputer Centre.
{
    \small
    \bibliographystyle{ieeenat_fullname}
    \bibliography{main}

@String(IJCV = {Int. J. Comput. Vis.})

@String(CVPR= {IEEE Conf. Comput. Vis. Pattern Recog.})

@String(ICCV= {Int. Conf. Comput. Vis.})

@String(ECCV= {Eur. Conf. Comput. Vis.})

@String(ICLR = {Int. Conf. Learn. Represent.})

@String(CVPRW= {IEEE Conf. Comput. Vis. Pattern Recog. Worksh.})

@String(IJCV  = {IJCV})

@String(CVPR  = {CVPR})

@String(ICCV  = {ICCV})

@String(ECCV  = {ECCV})

@String(ICLR  = {ICLR})

@String(CVPRW= {CVPRW})

@misc{ash,
      title={Extremely Simple Activation Shaping for Out-of-Distribution Detection}, 
      booktitle={ICLR},
      author={Andrija Djurisic and Nebojsa Bozanic and Arjun Ashok and Rosanne Liu},
      year={2023}
}

@inproceedings{
xu2024scaling,
title={Scaling for Training Time and Post-hoc Out-of-distribution Detection Enhancement},
author={Kai Xu and Rongyu Chen and Gianni Franchi and Angela Yao},
booktitle={ICLR},
year={2024},
}

@inproceedings{guo2017calibration,
  title={On calibration of modern neural networks},
  author={Guo, Chuan and Pleiss, Geoff and Sun, Yu and Weinberger, Kilian Q},
  booktitle={ICML},
  year={2017},
}

@article{kamoi2020mahalanobis,
  title={Why is the mahalanobis distance effective for anomaly detection?},
  author={Kamoi, Ryo and Kobayashi, Kei},
  journal={arXiv preprint arXiv:2003.00402},
  year={2020}
}

@inproceedings{du2022vos,
title={VOS: Learning What You Don't Know by Virtual Outlier Synthesis},
author={Xuefeng Du and Zhaoning Wang and Mu Cai and Sharon Li},
booktitle={ICLR},
year={2022},
 
}

@article{ming2022cider,
  title={CIDER: Exploiting Hyperspherical Embeddings for Out-of-Distribution Detection},
  author={Ming, Yifei and Sun, Yiyou and Dia, Ousmane and Li, Yixuan},
  journal={arXiv preprint arXiv:2203.04450},
  year={2022}
}

@article{Image-net-1k,
Author = {Olga Russakovsky and Jia Deng and Hao Su and Jonathan Krause and Sanjeev Satheesh and Sean Ma and Zhiheng Huang and Andrej Karpathy and Aditya Khosla and Michael Bernstein and Alexander C. Berg and Li Fei-Fei},
Title = {{ImageNet Large Scale Visual Recognition Challenge}},
Year = {2015},
journal   = { IJCV},
}

@inproceedings{iNaturalist,
  author={Van Horn, Grant and Mac Aodha, Oisin and Song, Yang and Cui, Yin and Sun, Chen and Shepard, Alex and Adam, Hartwig and Perona, Pietro and Belongie, Serge},
  booktitle={CVPR}, 
  title={The iNaturalist Species Classification and Detection Dataset}, 
  year={2018},
   
}

@inproceedings{textures,
  author={Cimpoi, Mircea and Maji, Subhransu and Kokkinos, Iasonas and Mohamed, Sammy and Vedaldi, Andrea},
  booktitle={CVPR}, 
  title={Describing Textures in the Wild}, 
  year={2014},
  
  }

@inproceedings{DanMSP,
title={A Baseline for Detecting Misclassified and Out-of-Distribution Examples in Neural Networks},
author={Dan Hendrycks and Kevin Gimpel},
booktitle={ICLR},
year={2017},
}

@inproceedings{ODIN,
title={Enhancing The Reliability of Out-of-distribution Image Detection in Neural Networks},
author={Shiyu Liang and Yixuan Li and R. Srikant},
booktitle={ICLR},
year={2018},
}

@inproceedings{Maha,
 author = {Lee, Kimin and Lee, Kibok and Lee, Honglak and Shin, Jinwoo},
 booktitle = {NeurIPS},
 title = {A Simple Unified Framework for Detecting Out-of-Distribution Samples and Adversarial Attacks},
 year = {2018}
}

@inproceedings{energybased_ood,
 author = {Liu, Weitang and Wang, Xiaoyun and Owens, John and Li, Yixuan},
 booktitle ={NeurIPS},
 title = {Energy-based Out-of-distribution Detection},
 year = {2020}
}

@inproceedings{nn-guide,
  title={Nearest Neighbor Guidance for Out-of-Distribution Detection},
  author={Park, Jaewoo and Jung, Yoon Gyo and Teoh, Andrew Beng Jin},
  booktitle={ICCV},
  year={2023}
}

@inproceedings{logitnorm,
title={Mitigating Neural Network Overconfidence with Logit Normalization},
author={Wei, Hongxin and Xie, Renchunzi and Cheng, Hao and Feng, Lei and An, Bo and Li, Yixuan},
booktitle={ICML},
year={2022}
}

@inproceedings{ViM,
  author={Wang, Haoqi and Li, Zhizhong and Feng, Litong and Zhang, Wayne},
  booktitle={CVPR}, 
  title={ViM: Out-Of-Distribution with Virtual-logit Matching}, 
  year={2022},
  
}

@inproceedings{
neco,
title={{NECO}: {NE}ural Collapse Based Out-of-distribution detection},
author={Mou{\"\i}n Ben Ammar and Nacim Belkhir and Sebastian Popescu and Antoine Manzanera and Gianni Franchi},
booktitle={ICLR},
year={2024},
}

@inproceedings{du2023dream,
      title={Dream the Impossible: Outlier Imagination with Diffusion Models}, 
      author={Xuefeng Du and Yiyou Sun and Xiaojin Zhu and Yixuan Li },
      booktitle={NeurIPS},
      year = {2023}
}

@article{csi,
  title={Csi: Novelty detection via contrastive learning on distributionally shifted instances},
  author={Tack, Jihoon and Mo, Sangwoo and Jeong, Jongheon and Shin, Jinwoo},
  journal={NeurIPS},
  year={2020}
}

@inproceedings{ding2025revisiting,
  title={Revisiting likelihood-based out-of-distribution detection by modeling representations},
  author={Ding, Yifan and Aleksandraus, Arturas and Ahmadian, Amirhossein and Unger, Jonas and Lindsten, Fredrik and Eilertsen, Gabriel},
  booktitle={SCIA},
  year={2025},
  organization={Springer}
}

@inproceedings{ahmadian2024unsupervised,
  title={Unsupervised novelty detection in pretrained representation space with locally adapted likelihood ratio},
  author={Ahmadian, Amirhossein and Ding, Yifan and Eilertsen, Gabriel and Lindsten, Fredrik},
  booktitle={AISTATS},
  year={2024},
}

@inproceedings{bai2023feed,
  title        = {Feed Two Birds with One Scone: Exploiting Wild Data for Both Out-of-Distribution Generalization and Detection},
  author       = {Bai, Haoyue and Canal, Gregory and Du, Xuefeng and Kwon, Jeongyeol and Nowak, Robert D. and Li, Yixuan},
  booktitle    = {ICML},
  year         = {2023}
}

@inproceedings{bai2024aha,
  title        = {AHA: Adaptive Human-Assisted Out-of-Distribution Generalization and Detection},
  author       = {Bai, Haoyue and Zhang, Jifan and Nowak, Robert},
  booktitle    = {NeurIPS},
  year         = {2024}
}

@article{zhu2025infobound,
  title        = {InfoBound: A Provable Information-Bounds Inspired Framework for Both OoD Generalization and OoD Detection},
  author       = {Zhu, Lin and Yang, Yifeng and Nie, Zichao and Gao, Yuan and Li, Jiarui and Gu, Qinying and Wang, Xinbing and Zhou, Chenghu and Ye, Nanyang},
  journal      = {TPAMI},
  year         = {2025}
}

@inproceedings{pereyra2017regularizing,
      title={Regularizing Neural Networks by Penalizing Confident Output Distributions}, 
      author={Gabriel Pereyra and George Tucker and Jan Chorowski and Łukasz Kaiser and Geoffrey Hinton},
      year={2017},
      booktitle={ICLR}
}

@inproceedings{
xiong2023proximityinformed,
title={Proximity-Informed Calibration for Deep Neural Networks},
author={Miao Xiong and Ailin Deng and Pang Wei Koh and Jiaying Wu and Shen Li and Jianqing Xu and Bryan Hooi},
booktitle={NeurIPS},
year={2023},
 
}

@inproceedings{PTS,
      title={Parameterized Temperature Scaling for Boosting the Expressive Power in Post-Hoc Uncertainty Calibration}, 
      author={Christian Tomani and Daniel Cremers and Florian Buettner},
      year={2022},
    booktitle={ECCV}
}

@InProceedings{mixnmatch,
  title = 	 {Mix-n-Match : Ensemble and Compositional Methods for Uncertainty Calibration in Deep Learning},
  author =       {Zhang, Jize and Kailkhura, Bhavya and Han, T. Yong-Jin},
  booktitle = { International Conference on Machine Learning},
  year = 	 {2020},
}

@inproceedings{histogram_binning,
author = {Zadrozny, Bianca and Elkan, Charles},
title = {Obtaining Calibrated Probability Estimates from Decision Trees and Naive Bayesian Classifiers},
booktitle = {ICML},
year = {2001}
}

@inproceedings{MI_binning,
title={Multi-Class Uncertainty Calibration via Mutual Information Maximization-based Binning},
author={Kanil Patel and William H. Beluch and Bin Yang and Michael Pfeiffer and Dan Zhang},
booktitle={ICLR},
year={2021},
 
}

@inproceedings{npos,
  title={Non-parametric Outlier Synthesis},
  author={Leitian Tao and Xuefeng Du and Jerry Zhu and Yixuan Li},
  booktitle={ICLR},
  year={2023},
  url={https://openreview.net/forum?id=JHklpEZqduQ}
}

@article{cider,
  title={CIDER: Exploiting Hyperspherical Embeddings for Out-of-Distribution Detection},
  author={Ming, Yifei and Sun, Yiyou and Dia, Ousmane and Li, Yixuan},
  journal={ arXiv:2203.04450},
  year={2022}
}

@article{mcinnes2018umap,
  title={Umap: Uniform manifold approximation and projection for dimension reduction},
  author={McInnes, Leland and Healy, John and Melville, James},
  journal={arXiv preprint arXiv:1802.03426},
  year={2018}
}

@article{mucsanyi2024benchmarking,
  title={Benchmarking uncertainty disentanglement: Specialized uncertainties for specialized tasks},
  author={Mucs{\'a}nyi, B{\'a}lint and Kirchhof, Michael and Oh, Seong Joon},
  journal={NeurIPS},
  year={2024}
}

@article{van2008visualizing,
  title={Visualizing data using t-SNE.},
  author={Van der Maaten, Laurens and Hinton, Geoffrey},
  journal={JMLR},
  year={2008}
}

@inproceedings{KL-matching,
  title = 	 {Scaling Out-of-Distribution Detection for Real-World Settings},
  author =       {Hendrycks, Dan and Basart, Steven and Mazeika, Mantas and Zou, Andy and Kwon, Joseph and Mostajabi, Mohammadreza and Steinhardt, Jacob and Song, Dawn},
  booktitle = {ICML},
   
  year = 	 {2022},
   
}

@inproceedings{sun2022knnood,
  title={Out-of-distribution Detection with Deep Nearest Neighbors},
  author={Sun, Yiyou and Ming, Yifei and Zhu, Xiaojin and Li, Yixuan},
  booktitle={ICML},
  year={2022}
}

@inproceedings{stable_diffusion,
      title={High-Resolution Image Synthesis with Latent Diffusion Models}, 
      author={Robin Rombach and Andreas Blattmann and Dominik Lorenz and Patrick Esser and Björn Ommer},
      year={2021},
    booktitle= {CVPR}
}

@inproceedings{zhang2023openood,
      title={OpenOOD v1.5: Enhanced Benchmark for Out-of-Distribution Detection}, 
      author={Jingyang Zhang and Jingkang Yang and Pengyun Wang and Haoqi Wang and Yueqian Lin and Haoran Zhang and Yiyou Sun and Xuefeng Du and Kaiyang Zhou and Wayne Zhang and Yixuan Li and Ziwei Liu and Yiran Chen and Hai Li},
      year={2024},
      booktitle={DMLR}
}

@inproceedings{ReAct,
  title={ReAct: Out-of-distribution Detection With Rectified Activations},
  author={Sun, Yiyou and Guo, Chuan and Li, Yixuan},
  booktitle={NeurIPS},
  year={2021}
}

@inproceedings{rank,
  title={RankFeat: Rank-1 Feature Removal for Out-of-distribution Detection},
  author={Song, Yue and Sebe, Nicu and Wang, Wei},
  booktitle = {NeurIPS},
  year={2022}
}

@inproceedings{Generalized_ODIN,
   
  author = {Hsu, Yen-Chang and Shen, Yilin and Jin, Hongxia and Kira, Zsolt},
  title = {Generalized ODIN: Detecting Out-of-distribution Image without Learning from Out-of-distribution Data},
  booktitle = {CVPR},
  year = {2020},
}

@inproceedings{outlier_exposure_for_DNN,
    title={Deep Anomaly Detection with Outlier Exposure},
    author={Dan Hendrycks and Mantas Mazeika and Thomas Dietterich},
    booktitle={ICLR},
    year={2019},
     
}

@inproceedings{isotonic_regression,
author = {Zadrozny, Bianca and Elkan, Charles},
title = {Transforming Classifier Scores into Accurate Multiclass Probability Estimates},
year = {2002},
booktitle = {ACM SIGKDD},
}

@InProceedings{calibration_modern_nn,
title = 	 {On Calibration of Modern Neural Networks},
author =       {Chuan Guo and Geoff Pleiss and Yu Sun and Kilian Q. Weinberger},
booktitle = 	 {ICML},
year = 	 {2017},
}

@inproceedings{xixi2023GEN,
title = {GEN: Pushing the Limits of Softmax-Based Out-of-Distribution Detection},
author = {Liu, Xixi and Lochman, Yaroslava and  Zach, Christopher},
booktitle = {CVPR},
year = {2023}
}

@article{cifar,
  title={Learning multiple layers of features from tiny images},
  author={Krizhevsky, A. and Hinton, G.},
  journal={Master's thesis, Department of Computer Science, University of Toronto},
  year={2009},
  publisher={Citeseer}
}

@inproceedings{svhn,
    author = {Netzer, Yuval and Wang, Tao and Coates, Adam and Bissacco, Alessandro and Wu, Bo and Ng, Andrew},
    title = {Reading Digits in Natural Images with Unsupervised Feature Learning},
    booktitle = {NeurIPS},
    year = {2011}
}

@article{liu2023fast,
  title={Fast decision boundary based out-of-distribution detector},
  author={Liu, Litian and Qin, Yao},
  journal={arXiv preprint arXiv:2312.11536},
  year={2023}
}

@INPROCEEDINGS {t2fnorm,
author = { Regmi, Sudarshan and Panthi, Bibek and Dotel, Sakar and Gyawali, Prashnna K and Stoyanov, Danail and Bhattarai, Binod },
booktitle = { CVPRW },
title = {{T2FNorm: Train-time Feature Normalization for OOD Detection in Image Classification}},
year = {2024}
}

@inproceedings{park2023understanding,
  title={Understanding the feature norm for out-of-distribution detection},
  author={Park, Jaewoo and Chai, Jacky Chen Long and Yoon, Jaeho and Teoh, Andrew Beng Jin},
  booktitle={CVPR},
  year={2023}
}

@inproceedings{yu2023block,
  title={Block selection method for using feature norm in out-of-distribution detection},
  author={Yu, Yeonguk and Shin, Sungho and Lee, Seongju and Jun, Changhyun and Lee, Kyoobin},
  booktitle={CVPR},
  year={2023}
}

@inproceedings{hua2021feature,
  title={On feature decorrelation in self-supervised learning},
  author={Hua, Tianyu and Wang, Wenxiao and Xue, Zihui and Ren, Sucheng and Wang, Yue and Zhao, Hang},
  booktitle={ICCV},
  year={2021}
}

@inproceedings{zhu2025addressing,
  title={Addressing representation collapse in vector quantized models with one linear layer},
  author={Zhu, Yongxin and Li, Bocheng and Xin, Yifei and Xia, Zhihua and Xu, Linli},
  booktitle={ICCV},
  year={2025}
}

@inproceedings{arjovsky2017wasserstein,
  title={Wasserstein generative adversarial networks},
  author={Arjovsky, Martin and Chintala, Soumith and Bottou, L{\'e}on},
  booktitle={ICML},
  year={2017},
}

@article{miyato2018spectral,
  title={Spectral normalization for generative adversarial networks},
  author={Miyato, Takeru and Kataoka, Toshiki and Koyama, Masanori and Yoshida, Yuichi},
  journal={arXiv preprint arXiv:1802.05957},
  year={2018}
}

@article{jing2021understanding,
  title={Understanding dimensional collapse in contrastive self-supervised learning},
  author={Jing, Li and Vincent, Pascal and LeCun, Yann and Tian, Yuandong},
  journal={arXiv preprint arXiv:2110.09348},
  year={2021}
}

@inproceedings{he2022exploring,
  title={Exploring the gap between collapsed \& whitened features in self-supervised learning},
  author={He, Bobby and Ozay, Mete},
  booktitle={ICML},
  year={2022},
}

@article{ghosh2022investigating,
  title={Investigating power laws in deep representation learning},
  author={Ghosh, Arna and Mondal, Arnab Kumar and Agrawal, Kumar Krishna and Richards, Blake},
  journal={arXiv preprint arXiv:2202.05808},
  year={2022}
}

@inproceedings{li2022understanding,
  title={Understanding collapse in non-contrastive siamese representation learning},
  author={Li, Alexander C and Efros, Alexei A and Pathak, Deepak},
  booktitle={ECCV},
  year={2022},
}

@inproceedings{garrido2023rankme,
  title={Rankme: Assessing the downstream performance of pretrained self-supervised representations by their rank},
  author={Garrido, Quentin and Balestriero, Randall and Najman, Laurent and Lecun, Yann},
  booktitle={ICML},
  year={2023},
}
}
%\input{sec/appendix}

\end{document}

\title{Enhancing Out-of-Distribution Detection with Extended Logit Normalization \\ 
Supplementary Material}

\maketitle
\thispagestyle{empty}

%%%%%%%%%%%%%%%%%%%%%%%%%%%%%%%%%%%%%%%%%%%%%%%%%%%%%%%%%%%%
\section{Proof of Proposition 3.1}
\label{sec:prooffz}

Using singular value decomposition, we express the weight matrix as
\[
\mathbf{W}^\top \mathbf{z} = \mathbf{U} \mathbf{\Sigma} \mathbf{V}^\top \mathbf{z},
\]
where \( \mathbf{U} \) and \( \mathbf{V} \) are orthonormal matrices, and \( \mathbf{\Sigma} \) is a diagonal matrix containing the singular values \( \sigma_1, \dots, \sigma_{c} \). Since orthonormal transformations preserve norms, it follows that
\[
\|\mathbf{W}^\top \mathbf{z}\| = \|\mathbf{\Sigma} \mathbf{V}^\top \mathbf{z}\|.
\]
Expanding this expression, we obtain
\[
\|\mathbf{W}^\top \mathbf{z}\| = \sqrt{\sum_{i=1}^{d} \sigma_i^2 (z'_i)^2},
\]
where \( \mathbf{z}' = \mathbf{V}^\top \mathbf{z} \) represents the transformed feature vector. Applying the extremal singular values, we establish the following bound:
\[
\sigma_{\min} \|\mathbf{z}\| \leq \|\mathbf{W}^\top \mathbf{z}\| \leq \sigma_{\max} \|\mathbf{z}\|.
\]
Including the bias term \( \mathbf{b} \), we obtain
\[
\sigma_{\min} \|\mathbf{z}\| - \|\mathbf{b}\| \leq \|\mathbf{f}\| \leq \sigma_{\max} \|\mathbf{z}\| + \|\mathbf{b}\|.
\]
If the singular values cluster around their mean \( \bar{\sigma} \), we approximate
\[
\|\mathbf{f}\| \approx \bar{\sigma} \|\mathbf{z}\| + \eta,
\]
where \( \eta \) represents noise due to \( \mathbf{b} \) and singular value variations. In the special case where \( \|\mathbf{b}\| = 0 \) and \( \mathbf{W} \) is isotropic (i.e., all singular values are identical), strict proportionality holds:
\[
\|\mathbf{f}\| = \bar{\sigma} \|\mathbf{z}\|.
\]
Thus, Proposition~3.1 is proved.

%%%%%%%%%%%%%%%%%%%%%%%%%%%%%%%%%%%%%%%%%%%%%%%%%%%%%%%%%%%%
\section{Proof of Proposition 3.2}
\label{sec:proofspace}

The decision boundary between class \( z_{\max} \) and class \( i \) is defined as
\[
H_{i~z_{\max}} = \{ \mathbf{z} \mid (\mathbf{w}_{z_{\max}} - \mathbf{w}_i)^\top \mathbf{z} + (b_{z_{\max}} - b_i) = 0 \}, \quad \forall i \neq z_{\max}.
\]
If \( m \geq c - 1 \), there exists at least one solution, ensuring that \( \mathcal{D}_{\min}(\mathbf{z}) = 0 \). Their intersection forms an affine subspace of dimension
\[
\dim\left(\bigcap_{i \neq z_{\max}} H_{i~z_{\max}}\right) = m - c + 1.
\]
If \( m < c - 1 \), the linear system is overdetermined and admits no exact solution. Instead, we solve the least-squares problem
\[
\mathbf{A} \mathbf{z} = \mathbf{b},
\]
where \( \mathbf{A} \) is a \( (c-1) \times m \) matrix whose rows are given by \( (\mathbf{w}_{z_{\max}} - \mathbf{w}_i)^\top \), \( \mathbf{b} \) is a \( (c-1) \)-dimensional vector with entries \( b_{z_{\max}} - b_i \), and \( \mathbf{z} \) is the unknown \( m \)-dimensional vector. The approximate solution is given by
\[
\mathbf{z}^* = (\mathbf{A}^\top \mathbf{A})^{-1} \mathbf{A}^\top \mathbf{b}.
\]
Assume \( \mathbf{A}^\top \mathbf{A} \) (independent decision boundaries) is an invertible \( m \times m \) matrix. Then the solution space of \( \mathbf{z}^* \) is zero-dimensional, and
\[
\mathcal{D}_{\min}(\mathbf{z}) = \| \mathbf{A} \mathbf{z}^* - \mathbf{b} \|_2 > 0.
\]
Thus, Proposition~3.2 is proved.

%%%%%%%%%%%%%%%%%%%%%%%%%%%%%%%%%%%%%%%%%%%%%%%%%%%%%%%%%%%%
\section{Computation Overhead Analysis}
Our method introduces an additional pairwise weight-distance computation with theoretical complexity $\mathcal{O}(C^{2}D)$; the corresponding PyTorch implementation is shown in Fig.~\ref{fig:elogitnorm_code}. Although this appears costly, the overhead is negligible in practice. The key reason is that the pairwise operation is applied only to the final linear classifier rather than the backbone. Even for ImageNet-1K, where $C=1000$ and a typical classifier dimension is $D\approx 2048$, the resulting parameter tensor ($\sim$2M weights) is extremely small compared to the feature maps processed by modern CNN backbones. The entire pairwise norm computation is executed as a single dense, fully vectorized GPU kernel, requires no gradient backpropagation, and adds only a few milliseconds per iteration. Consequently, the overall training cost continues to be dominated by the forward--backward computation of the backbone network.

\noindent Empirically, we observe almost no additional wall-time when training on ImageNet-1K. For example, with ResNet-18 on ImageNet-1K, the extra classifier operation amounts to $C^{2}D = 1000^{2}\!\times\!512 \approx 5.12\times10^{8}$ FLOPs ($\approx 0.512$ GFLOPs), compared to the $\approx 5.4$ GFLOPs required for a forward--backward pass through the backbone---an overhead of only about 9--10\%. On smaller-scale datasets, the overhead becomes even more negligible. For ResNet-18 on CIFAR-100 ($C=100, D=512$), the additional work is only $100^{2}\!\times\!512 = 5.12\times10^{6}$ FLOPs ($\approx 0.005$ GFLOPs), which is less than 1\% of the $\approx 0.9$ GFLOPs consumed by the forward--backward pass. Across all settings we tested, the end-to-end training time remains nearly identical to standard cross-entropy training despite the asymptotically higher complexity.

\begin{figure}[t]
\begin{minipage}{\linewidth}
\begin{lstlisting}
import torch
import torch.nn.functional as F

def elogitnorm_loss(logits, fc_weight, target):
    """
    ELogitNorm loss corresponding to the equations in Appendix.
    logits:    (N, C)
    fc_weight: (C, D)
    target:    (N,)
    """

    # Pairwise classifier-weight differences
    w_diff = fc_weight.unsqueeze(1) - fc_weight.unsqueeze(0)
    denom = torch.norm(w_diff, dim=2)
    denom.fill_diagonal_(1.0)

    # Maximum logit and predicted class
    values, nn_idx = logits.max(dim=1)

    # Logit gaps
    gaps = (logits - values.unsqueeze(1)).abs()

    # Instance-wise scaling factor
    scale = (gaps / denom[nn_idx]).mean(dim=1, keepdim=True)

    # ELogitNorm objective
    scaled_logits = logits / scale
    loss = F.cross_entropy(scaled_logits, target)

    return loss
\end{lstlisting}
\end{minipage}
\caption{PyTorch-style implementation of the proposed ELogitNorm training objective corresponding to Eqns.~(7)--(8).}
\label{fig:elogitnorm_code}
\end{figure}

%%%%%%%%%%%%%%%%%%%%%%%%%%%%%%%%%%%%%%%%%%%%%%%%%%%%%%%%%%%%
\section{Extra Experiments}
\label{sec:extraexp}

To further assess the generality and practical utility of the proposed ELogitNorm, we conduct additional experiments on both small-scale and large-scale OOD benchmarks. The results are summarized in Tables~\ref{tab:CIFAR100-main} and~\ref{tab:ImageNet200-main}. Across all post-hoc detection methods evaluated, including \MSP, \ReAct, \KNN, \GEN, \fDBD, and \SCALE, their ELogitNorm-enhanced variants (denoted with ``$*$'') consistently outperform their standard counterparts on both near-OOD and far-OOD datasets.

\noindent\textbf{Per-dataset results on CIFAR-100.}
Table~\ref{tab:CIFAR100-main} reports the per-dataset OOD performance when the in-distribution (ID) dataset is CIFAR-100 and the encoder is ResNet18. ELogitNorm provides substantial improvements in far-OOD scenarios (e.g., MNIST, SVHN, Textures, and Places365). In particular, \MSPTAG, \ReActTAG, and \GENTAG exhibit clear gains in AUROC and considerable reductions in FPR, demonstrating that the distance-aware normalization effectively sharpens confidence transitions around decision boundaries. These improvements occur without any architectural modifications or additional training-time regularizers, confirming the plug-in nature of our method.

\begin{table*}
\centering
\caption{\emph{Per-dataset performance of OOD detection methods and their ELogitNorm-enhanced variants (denoted with $*$).} 
The image encoder is ResNet18. The ID dataset is \textbf{CIFAR-100}.}  
\begin{adjustbox}{width=\linewidth,center}
\begin{tabular}{llcccccccccccccccc}
\toprule
& \multirow{2}{*}{OOD Method}   &  
\multicolumn{2}{c}{\textbf{CIFAR-10}} &
\multicolumn{2}{c}{\textbf{TIN}} &
\multicolumn{2}{c}{\textbf{Near-OOD}} &
\multicolumn{2}{c}{\textbf{MNIST}} & 
\multicolumn{2}{c}{\textbf{SVHN}} &
\multicolumn{2}{c}{\textbf{Textures}} &
\multicolumn{2}{c}{\textbf{Places365}} &
\multicolumn{2}{c}{\textbf{Far-OOD}} \\
& & AUROC & FPR & AUROC & FPR & AUROC & FPR & AUROC & FPR & AUROC &  FPR & AUROC &  FPR & AUROC &  FPR & AUROC &  FPR \\
\doublerule

& \MSP & 78.47 & 58.91 & 82.07 & 50.70 & 80.27 & 54.80 & 76.08 & 57.23 & 78.42 & 59.07 & 77.32 & 61.88 & 79.22 & 56.62 & 77.76 & 58.70 \\
\rowcolor{mygray} & \MSPTAG & 76.00 & 69.15 & 83.37 & 49.81 & 79.68 & 59.48 & 90.23 & 30.88 & 88.56 & 37.73 & 78.16 & 64.33 & 81.07 & 54.51 & 84.51 & 46.86 \\
\midrule
& \ReAct & 78.65 & 61.30 & 82.88 & 51.47 & 80.77 & 56.39 & 78.37 & 56.04 & 83.01 & 50.41 & 80.15 & 55.04 & 80.03 & 55.30 & 80.39 & 54.20 \\
\rowcolor{mygray} & \ReActTAG & 73.80 & 74.76 & 82.27 & 50.96 & 78.04 & 62.86 & 91.86 & 27.45 & 90.19 & 33.13 & 77.87 & 59.37 & 80.25 & 56.38 & 85.04 & 44.08 \\
\midrule
& \KNN & 77.02 & 72.80 & 83.34 & 49.65 & 80.18 & 61.22 & 82.36 & 48.58 & 84.15 & 51.75 & 83.66 & 53.56 & 79.43 & 60.70 & 82.40 & 53.65 \\
\rowcolor{mygray} & \KNNTAG & 77.11 & 65.94 & 83.14 & 50.26 & 80.12 & 58.10 & 85.93 & 44.80 & 83.89 & 56.63 & 82.78 & 55.01 & 78.74 & 60.48 & 82.84 & 54.23 \\
\midrule
& \GEN & 79.38 & 58.87 & 83.25 & 49.98 & 81.31 & 54.42 & 78.29 & 53.92 & 81.41 & 55.45 & 78.74 & 61.23 & 80.28 & 56.25 & 79.68 & 56.71 \\
\rowcolor{mygray} & \GENTAG & 75.20 & 69.42 & 82.75 & 50.18 & 78.98 & 59.80 & 91.50 & 29.31 & 89.29 & 36.51 & 76.88 & 64.73 & 80.74 & 54.71 & 84.60 & 46.32 \\
\midrule
& \fDBD & 78.35 & 63.89 & 83.97 & 47.89 & 81.16 & 55.89 & 79.05 & 51.35 & 80.48 & 53.80 & 81.18 & 53.65 & 79.85 & 57.16 & 80.14 & 53.99 \\
\rowcolor{mygray} & \fDBDTAG & 75.60 & 73.73 & 83.32 & 50.20 & 79.46 & 61.96 & 90.23 & 31.84 & 88.78 & 38.70 & 82.08 & 50.91 & 80.52 & 56.44 & 85.40 & 44.47 \\
\midrule
& \SCALE &79.26 & 59.11 & 82.71 & 52.24 & 80.99 & 55.68 & 80.27 & 51.64 & 84.45 & 49.27 & 80.50 & 58.45 & 80.47 & 56.98 & 81.42 & 54.09 \\
\rowcolor{mygray} & \SCALETAG & 74.99 & 69.91 & 82.73 & 50.36 & 78.86 & 60.14 & 91.72 & 28.60 & 89.68 & 35.58 & 77.46 & 64.02 & 80.82 & 54.56 & 84.92 & 45.69 \\
\bottomrule
\end{tabular}
\end{adjustbox}
\label{tab:CIFAR100-main}
\end{table*}

\noindent\textbf{Per-dataset results on ImageNet-200.}
Table~\ref{tab:ImageNet200-main} presents analogous results when the ID dataset is ImageNet-200. Here the label space is significantly larger and the OOD shifts are more challenging. ELogitNorm again consistently boosts performance for all tested post-hoc detectors. For example, \KNNTAG achieves strong performance on both iNaturalist and OpenImage-O, and \SCALETAG attains state-of-the-art far-OOD detection on ImageNet-200 benchmarks such as Textures and OpenImage-O. These results indicate that ELogitNorm scales favorably with the number of classes and benefits detectors that rely on logits, features, or activation statistics alike.
\begin{table*}[h] \centering \caption{\emph{Per-dataset performance of OOD detection methods and their ELogitNorm-enhanced variants (denoted with $*$).} The image encoder is ResNet18. The ID dataset is \textbf{ImageNet-200}.} \begin{adjustbox}{width=\linewidth,center} \begin{tabular}{llcccccccccccccc} \toprule & \multirow{2}{*}{OOD Method} & \multicolumn{2}{c}{\textbf{SSB-hard}} & \multicolumn{2}{c}{\textbf{NINCO}} & \multicolumn{2}{c}{\textbf{Near-OOD}} & \multicolumn{2}{c}{\textbf{iNaturalist}} & \multicolumn{2}{c}{\textbf{Textures}} & \multicolumn{2}{c}{\textbf{OpenImage-O}} & \multicolumn{2}{c}{\textbf{Far-OOD}} \\ & & AUROC & FPR & AUROC & FPR & AUROC & FPR & AUROC & FPR & AUROC & FPR & AUROC & FPR & AUROC & FPR \\ \doublerule & \MSP & 80.38 & 66.00 & 86.29 & 43.65 & 83.34 & 54.82 & 92.80 & 26.48 & 88.36 & 44.58 & 89.24 & 35.23 & 90.13 & 35.43 \\ \rowcolor{mygray} & \MSPTAG & 79.00 & 66.01 & 87.12 & 43.96 & 83.06 & 54.99 & 96.50 & 14.75 & 92.56 & 31.81 & 91.70 & 28.68 & 93.58 & 25.08 \\ \midrule & \ReAct & 78.97 & 71.51 & 84.76 & 53.47 & 81.87 & 62.49 & 93.65 & 22.97 & 92.86 & 29.67 & 90.40 & 32.86 & 92.31 & 28.50 \\ \rowcolor{mygray} & \ReActTAG & 73.24 & 75.78 & 83.87 & 54.66 & 78.56 & 65.22 & 96.27 & 16.89 & 92.91 & 32.55 & 90.10 & 35.97 & 93.10 & 28.47 \\ \midrule & \KNN & 77.03 & 73.71 & 86.10 & 46.64 & 81.57 & 60.18 & 93.99 & 24.46 & 95.29 & 24.45 & 90.19 & 32.90 & 93.16 & 27.27 \\ \rowcolor{mygray} & \KNNTAG & 78.18 & 69.64 & 87.27 & 43.66 & 82.73 & 56.65 & 97.17 & 12.86 & 97.00 & 16.79 & 94.06 & 24.47 & 96.08 & 18.04 \\ \midrule & \GEN & 80.75 & 66.79 & 86.60 & 43.61 & 83.68 & 55.20 & 93.70 & 22.03 & 90.25 & 42.01 & 90.13 & 32.25 & 91.36 & 32.10 \\ \rowcolor{mygray} & \GENTAG & 77.43 & 66.30 & 86.03 & 46.60 & 81.73 & 56.45 & 96.26 & 16.39 & 91.62 & 34.79 & 90.56 & 31.68 & 92.81 & 27.62 \\ \midrule & \fDBD & 78.62 & 72.42 & 87.78 & 42.11 & 83.20 & 57.27 & 96.84 & 14.19 & 93.95 & 26.37 & 92.56 & 26.59 & 94.45 & 22.38 \\ \rowcolor{mygray} & \fDBDTAG & 78.85 & 72.39 & 87.81 & 42.39 & 83.33 & 57.39 & 96.80 & 14.03 & 93.94 & 26.36 & 92.61 & 26.31 & 94.45 & 22.23 \\ \midrule & \SCALE & 82.08 & 67.39 & 87.60 & 47.20 & 84.84 & 57.29 & 95.79 & 18.41 & 94.11 & 29.75 & 92.04 & 31.21 & 93.98 & 26.46 \\ \rowcolor{mygray} & \SCALETAG & 75.73 & 71.33 & 85.95 & 48.86 & 80.84 & 60.10 & 97.09 & 13.64 & 95.85 & 20.70 & 92.68 & 28.88 & 95.21 & 21.08 \\ \bottomrule \end{tabular} \end{adjustbox} \label{tab:ImageNet200-main} \end{table*}

\begin{figure*}[h]
  \centering
  \begin{subfigure}{0.48\linewidth}
    \includegraphics[width=1\linewidth]{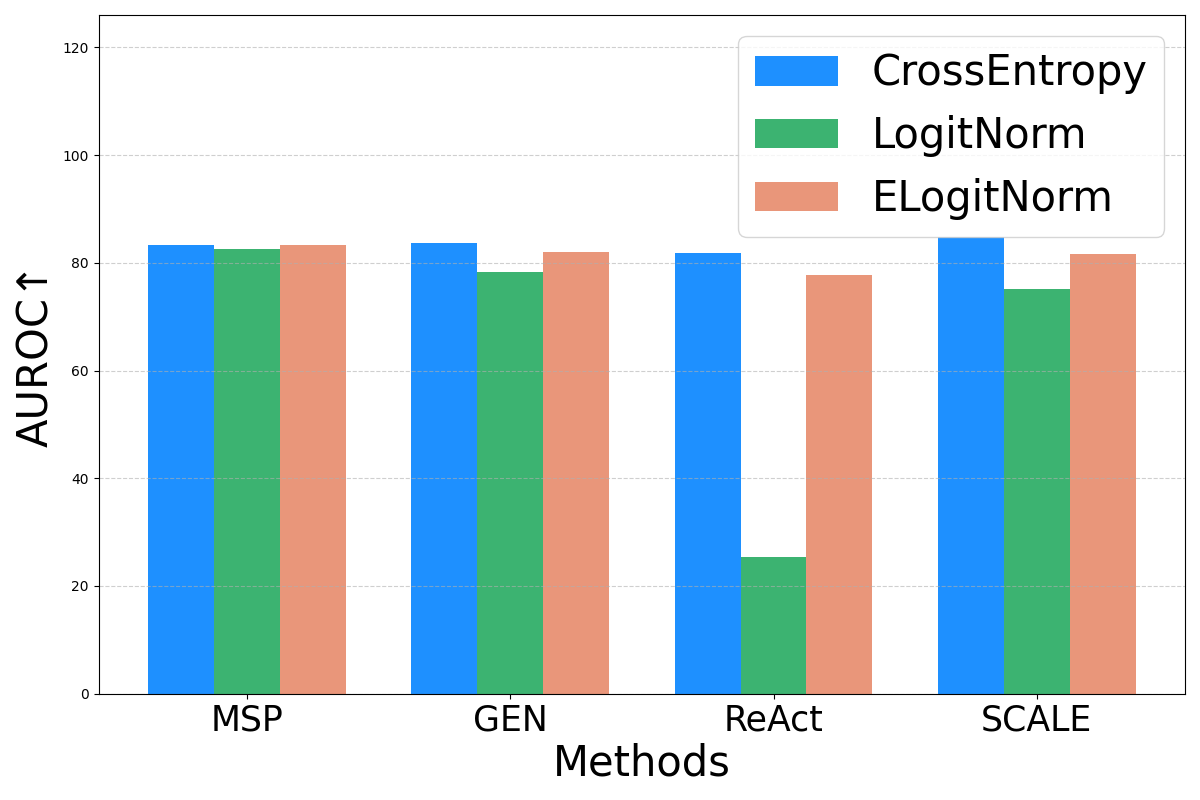}
    \caption{ImageNet-200, near-OOD}
    \label{fig:exa-a}
  \end{subfigure}
  \begin{subfigure}{0.48\linewidth}
    \includegraphics[width=1\linewidth]{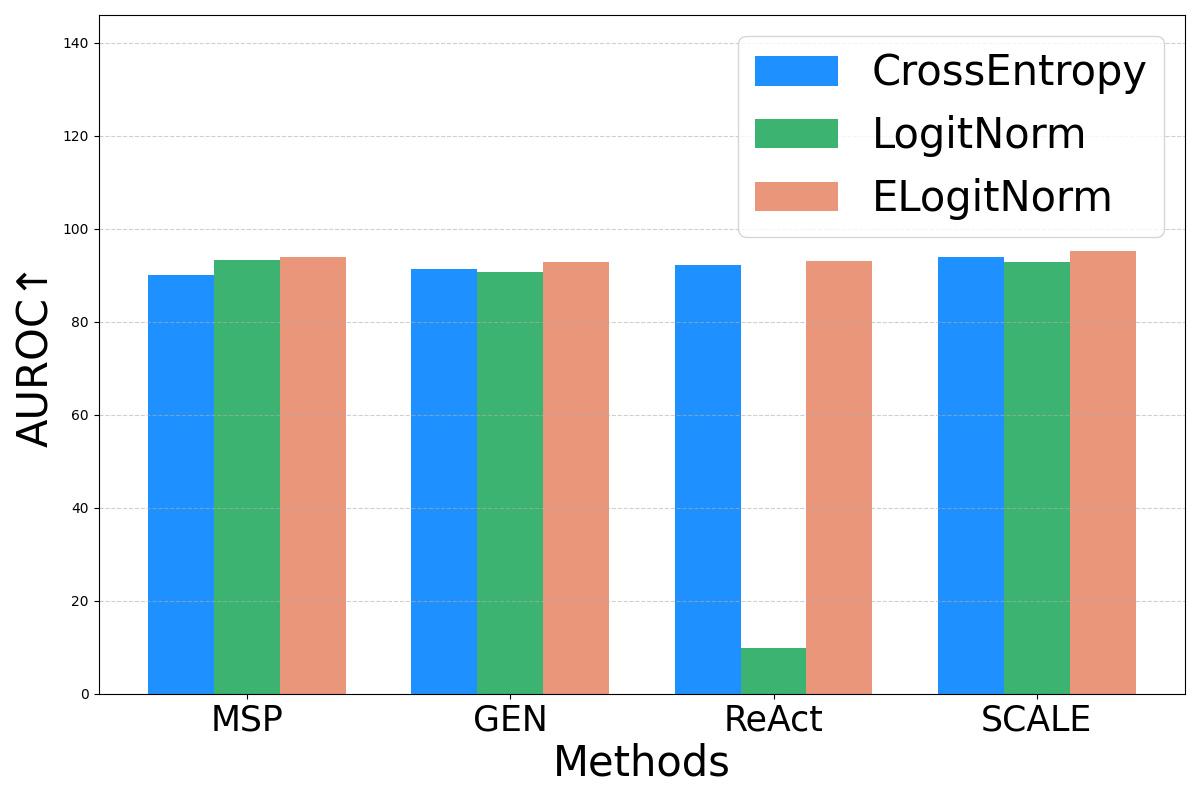}
    \caption{ImageNet-200, far-OOD}
    \label{fig:exa-b}
  \end{subfigure}
  \caption{\emph{Average near-OOD and far-OOD performance with four post-hoc methods \MSP, \ReAct, \GEN and \SCALE.} ResNet18 models are trained with \textcolor{eblue}{Cross-Entropy}, \textcolor{mygreen}{LogitNorm}~\cite{logitnorm}, and \textcolor{mysalmon}{\textbf{ELogitNorm} (Ours)}, respectively. ID data is ImageNet-200.}
  \label{fig:ex}
\end{figure*}

\begin{figure}
  \centering
  \begin{subfigure}{0.3\linewidth}
    \includegraphics[width=1\linewidth]{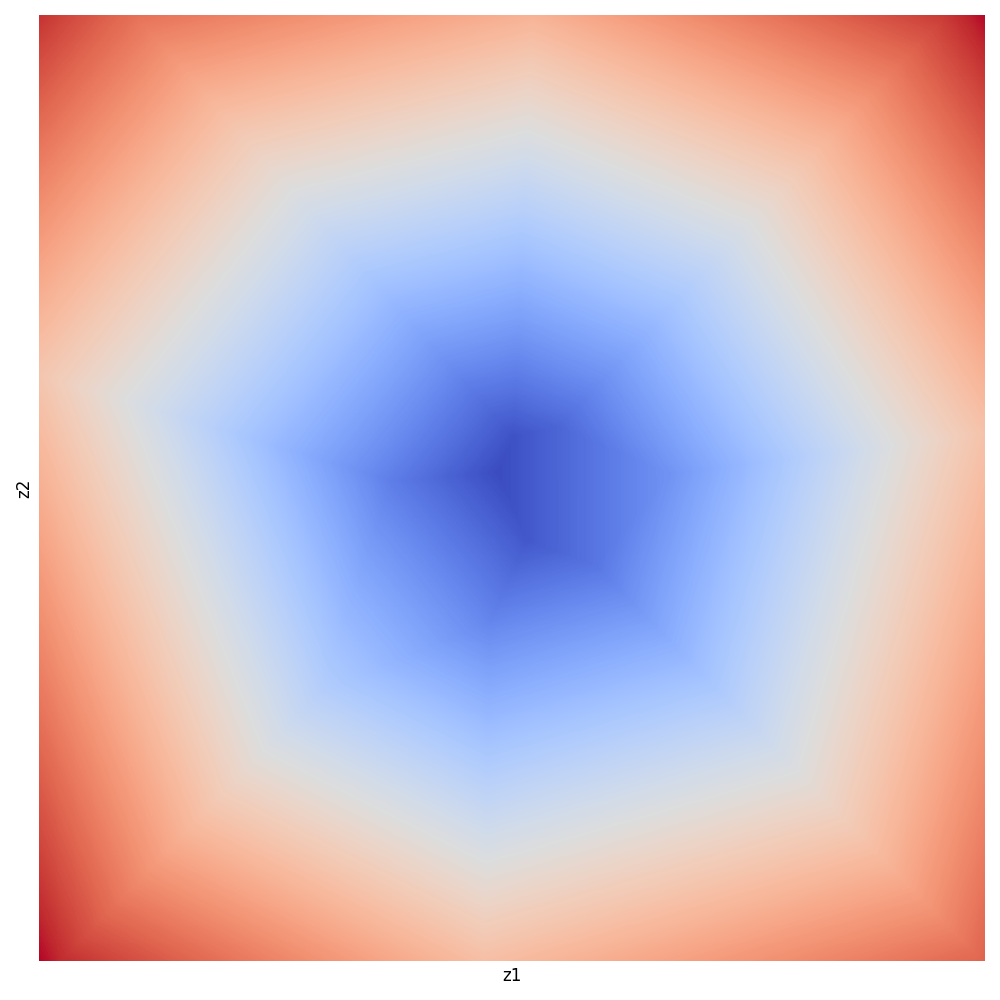}
    \caption{Cross-Entropy}
    \label{fig:ex-a}
  \end{subfigure}
  \begin{subfigure}{0.3\linewidth}
    \includegraphics[width=1\linewidth]{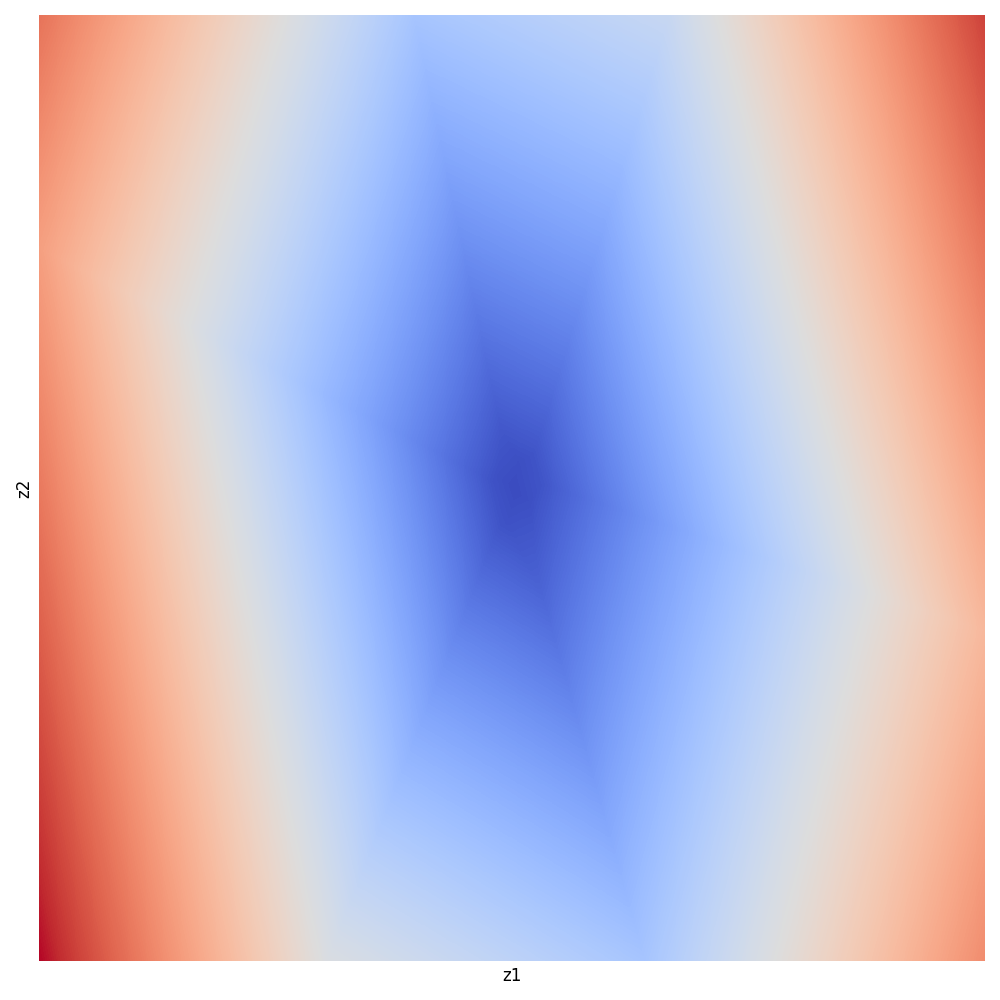}
    \caption{LogitNorm}
    \label{fig:ex-b}
  \end{subfigure}
  \begin{subfigure}{0.3\linewidth}
    \includegraphics[width=1\linewidth]{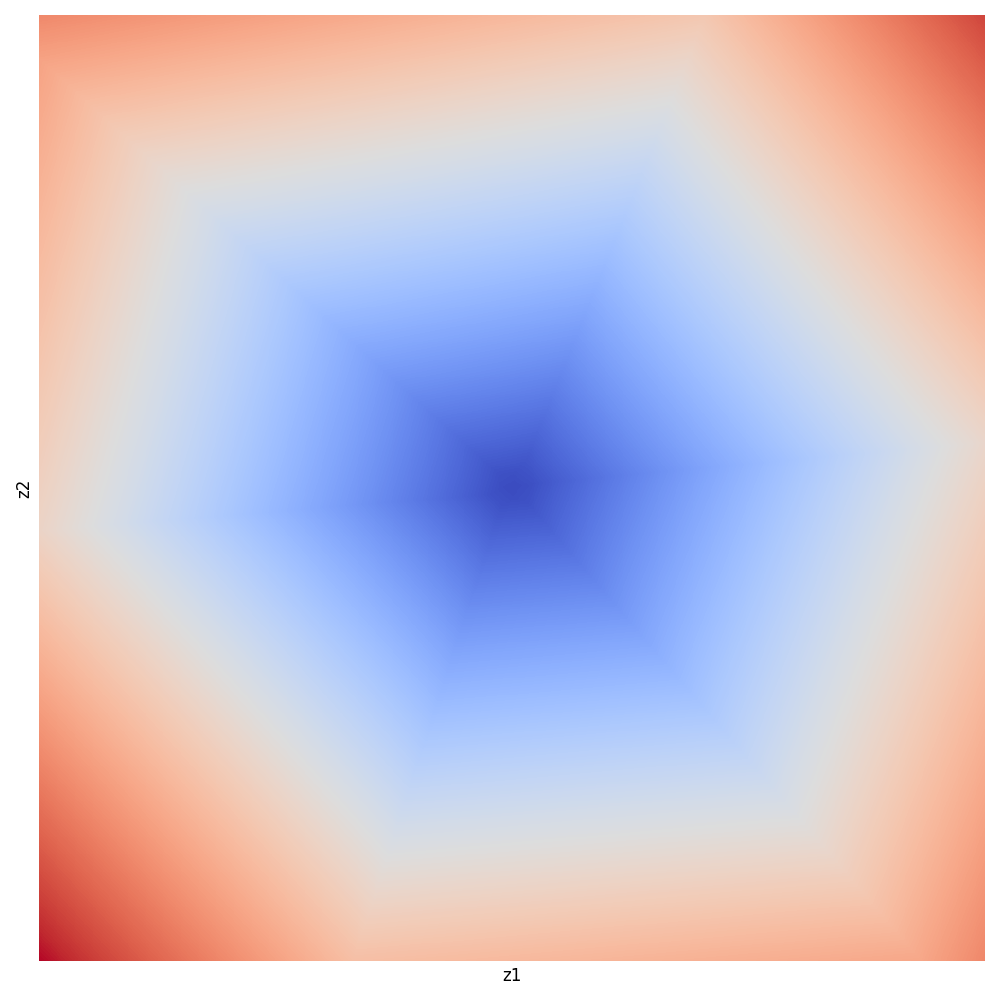}
    \caption{ELogitNorm}
    \label{fig:ex-c}
  \end{subfigure}
  \caption{Max-logit map on a 2D feature space, with the same setting as in Fig.~2(b). A ResNet18 is trained on CIFAR-10 and the feature is set to \(\mathbf{z} \in \mathbb{R}^2\) before the penultimate layer.}
  \label{fig:vis}
\end{figure}

\noindent\textbf{Max-logit visualization.}
Fig~\ref{fig:vis} visualizes the maximal logit over a 2D feature space for Cross-Entropy, LogitNorm, and ELogitNorm. Under this controlled setting, ELogitNorm produces smoother and more structured max-logit landscapes compared to the baselines. This geometric refinement translates into clearer decision regions and better separation between high- and low-confidence areas, which in turn supports more reliable OOD detection.

\begin{figure*}[h]
  \centering
  \begin{subfigure}{0.48\linewidth}
    \includegraphics[width=1\linewidth]{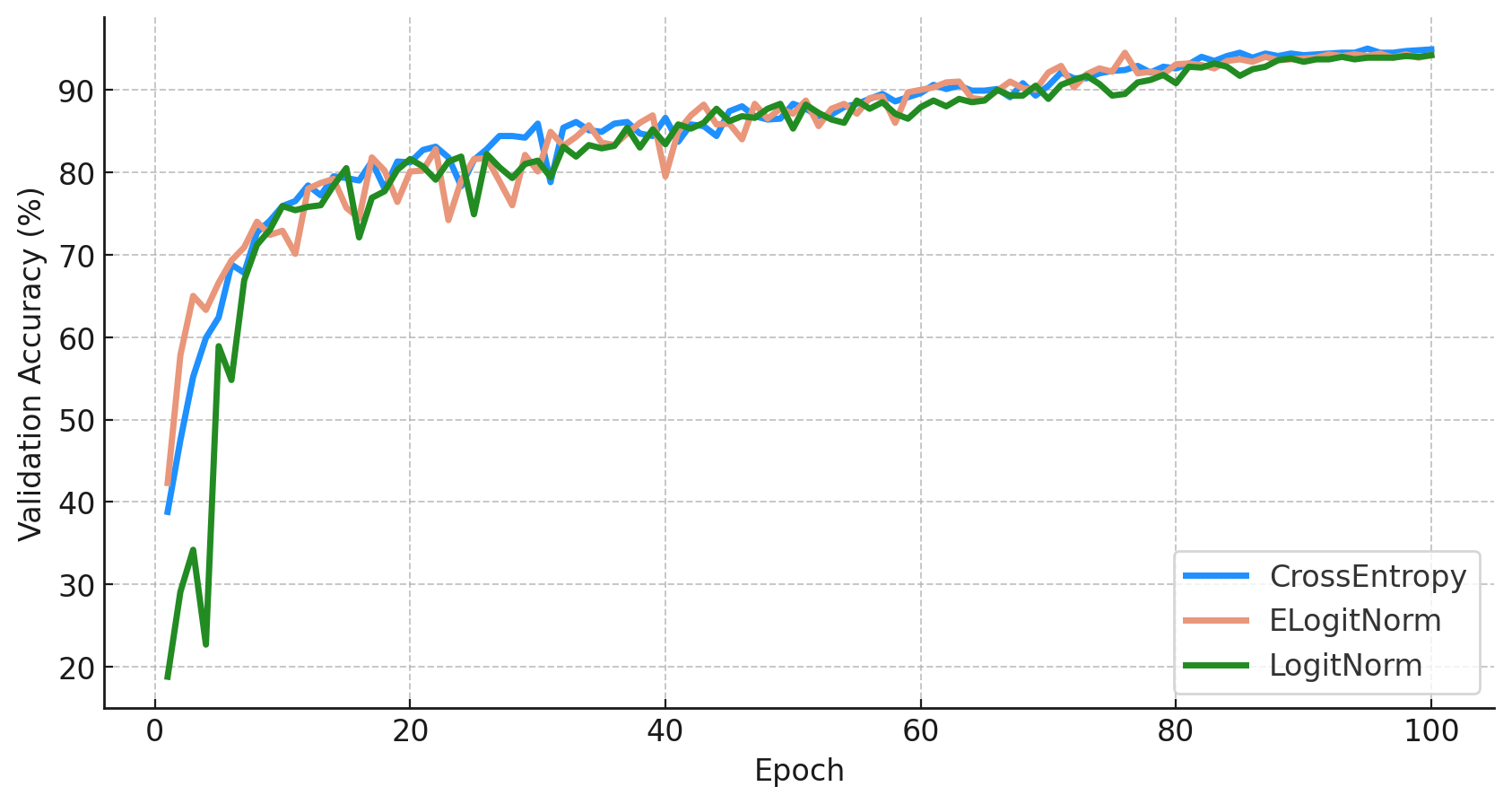}
    \caption{CIFAR-10}
    \label{fig:acc-a}
  \end{subfigure}
  \begin{subfigure}{0.48\linewidth}
    \includegraphics[width=1\linewidth]{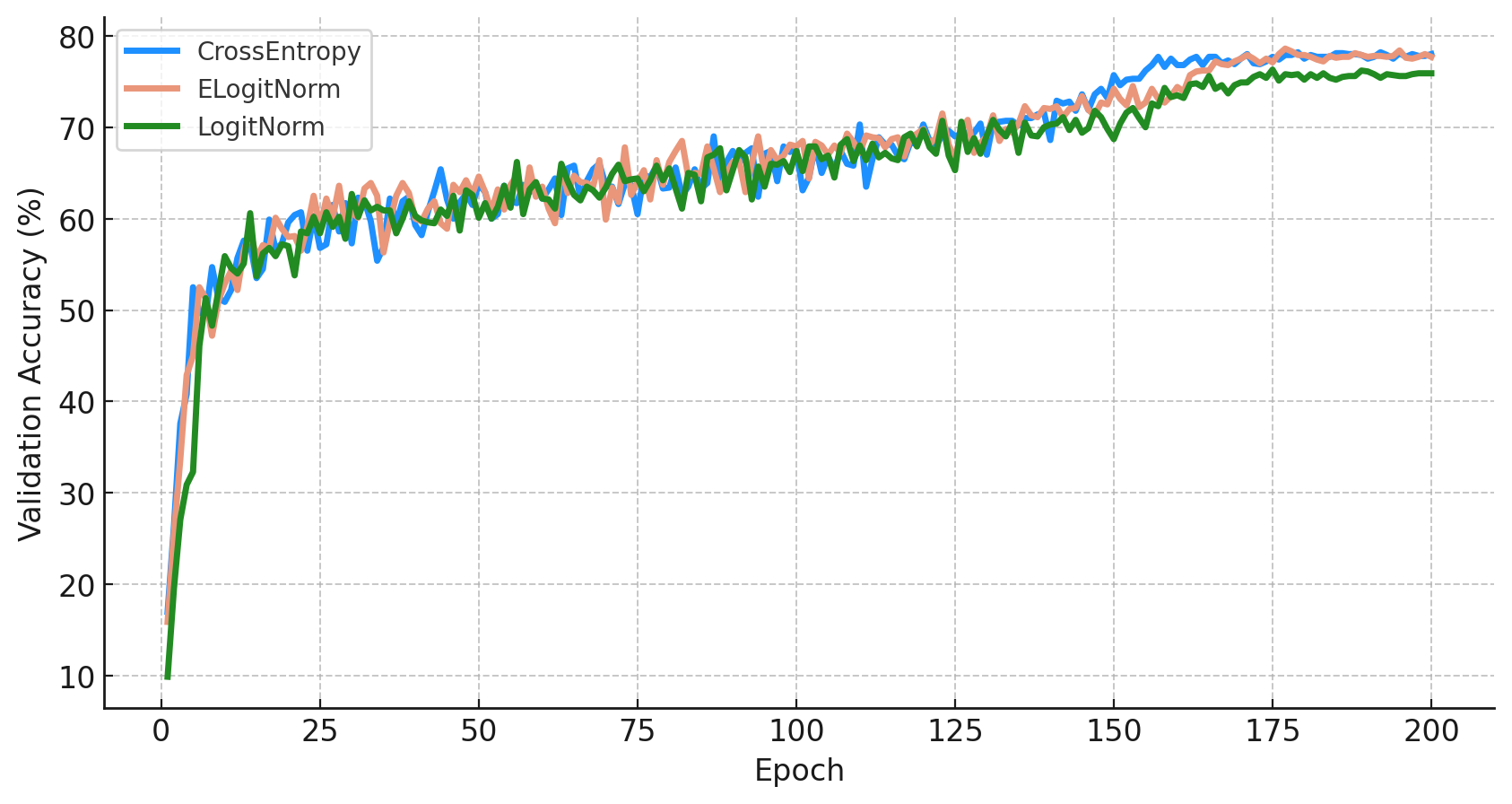}
    \caption{CIFAR-100}
    \label{fig:acc-b}
  \end{subfigure}
  \caption{\emph{Training stability.} ResNet18 models are trained on CIFAR-10 and CIFAR-100 with \textcolor{eblue}{Cross-Entropy}, \textcolor{mygreen}{LogitNorm}~\cite{logitnorm}, and \textcolor{mysalmon}{\textbf{ELogitNorm} (Ours)}, respectively.}
  \label{fig:acc}
\end{figure*}

\noindent\textbf{Training stability analysis.}
Fig~\ref{fig:acc} compares the training dynamics of ResNet18 on CIFAR-10 and CIFAR-100 using \textcolor{eblue}{Cross-Entropy}, \textcolor{mygreen}{LogitNorm}~\cite{logitnorm}, and \textcolor{mysalmon}{\textbf{ELogitNorm} (Ours)}. Across both datasets, ELogitNorm exhibits stable convergence behavior that closely matches the baselines, without introducing additional oscillations or slowdowns. On CIFAR-10, all three methods converge rapidly to similar final validation accuracy, with ELogitNorm showing smooth optimization despite the adaptive scaling applied to logits. On CIFAR-100, where optimization is typically more sensitive, ELogitNorm again tracks the trajectories of Cross-Entropy and LogitNorm, demonstrating reliable and consistent learning throughout the entire 200-epoch schedule. These results confirm that the instance-wise distance-aware normalization does not adversely affect the optimization landscape. Instead, ELogitNorm preserves the training characteristics of standard cross-entropy while offering significant improvements in OOD robustness, validating that the proposed normalization is both effective and practical to deploy in real training pipelines.

%%%%%%%%%%%%%%%%%%%%%%%%%%%%%%%%%%%%%%%%%%%%%%%%%%%%%%%%%%%%

%%%%%%%%% REFERENCES
{
    \small
    \bibliographystyle{ieeenat_fullname}
    \bibliography{main}
}